\documentclass[10pt,journal,compsoc]{IEEEtran}
\usepackage{graphicx}
\usepackage{capt-of}
\usepackage{amsmath,amssymb,amsfonts,dsfont,bm,bbm,mathrsfs,pifont}
\usepackage[ruled,linesnumbered]{algorithm2e}
\usepackage{listings}
\usepackage{booktabs,multirow,adjustbox}
\usepackage{float}
\usepackage[table]{xcolor}

\renewcommand{\paragraph}[1]{\vspace{1.25mm}\noindent\textbf{#1}}

\usepackage{hyperref}
\usepackage{graphicx}
\usepackage{color}
\definecolor{linkcolor}{RGB}{255,0,0}
\definecolor{urlcolor}{RGB}{255,105,180}
\definecolor{citecolor}{RGB}{66,168,235}
\definecolor{codegreen}{RGB}{57,182,74}
\hypersetup{colorlinks=true,linkcolor=linkcolor,urlcolor=urlcolor}
\newcommand{\ie}{\textit{i.e.}} 
\newcommand{\eg}{\textit{e.g.}} 
\usepackage{textcomp}
\usepackage{stfloats}
\usepackage{url}
\usepackage{verbatim}
\usepackage{pifont}
\newcommand{\cmark}{\ding{52}}%
\newcommand{\xmark}{\textcolor{gray}{\ding{56}}}%
\usepackage{graphicx}
\usepackage{booktabs} 
\usepackage[table]{xcolor}
\usepackage{multirow}
\definecolor{Light}{rgb}{0.99, 0.92, 0.95}

\newlength\savewidth
\definecolor{Light}{rgb}{0.99, 0.92, 0.95}

\ifCLASSOPTIONcompsoc
  \usepackage[nocompress]{cite}
\else
  \usepackage{cite}
\fi

\begin{document}

\title{INP-Former++:  Advancing Universal Anomaly Detection via Intrinsic Normal Prototypes\\ and Residual Learning}
%
%
%
%

\author{Wei~Luo,
        Haiming~Yao,
        Yunkang~Cao,
        Qiyu~Chen,
        Ang~Gao, \\ 
        Weiming~Shen,~\IEEEmembership{Fellow,~IEEE},
and~Wenyong~Yu$^*$,~\IEEEmembership{Senior~Member,~IEEE}
\IEEEcompsocitemizethanks{
\IEEEcompsocthanksitem 
Wei Luo, Haiming Yao, and Ang Gao are with the State Key Laboratory of
 Precision Measurement Technology and Instruments, Department of Precision Instrument, Tsinghua University, Beijing, China. (e-mail: \{luow23, yhm22, ga24\}@mails.tsinghua.edu.cn).
\IEEEcompsocthanksitem
Weiming Shen and Wenyong Yu are with the State Key Laboratory of Intelligent Manufacturing Equipment and Technology, Huazhong
 University of Science and Technology, Wuhan, China. (e-mail: wshen@ieee.org, ywy@hust.edu.cn).
\IEEEcompsocthanksitem 
Yunkang Cao is with Hunan University School of Robotics, Changsha, China. (e-mail: caoyunkang@ieee.org).
\IEEEcompsocthanksitem
Qiyu Chen is with the Institute of Automation, Chinese Academy of Sciences, Beijing, China. (e-mail: chenqiyu2021@ia.ac.cn).
}
\thanks{This study was supported by the National Natural Science Foundation of China under Grant No. 52375494. \textit{(Corresponding author: Wenyong Yu.)}}
}

%
%



\IEEEtitleabstractindextext{%
\begin{abstract}
Anomaly detection (AD) is essential for industrial inspection and medical diagnosis, yet existing methods typically rely on ``comparing'' test images to normal references from a training set. However, variations in appearance and positioning often complicate the alignment of these references with the test image, limiting detection accuracy. We observe that most anomalies manifest as local variations, meaning that even within anomalous images, valuable normal information remains. We argue that this information is useful and may be more aligned with the anomalies since both the anomalies and the normal information originate from the same image. Therefore, rather than relying on external normality from the training set, we propose INP-Former, a novel method that extracts \textit{Intrinsic Normal Prototypes (INPs)} directly from the test image. Specifically, we introduce the INP Extractor, which linearly combines normal tokens to represent INPs. We further propose an INP Coherence Loss to ensure INPs can faithfully represent normality for the testing image. These INPs then guide the INP-guided Decoder to reconstruct only normal tokens, with reconstruction errors serving as anomaly scores. Additionally, we propose a Soft Mining Loss to prioritize hard-to-optimize samples during training. INP-Former achieves state-of-the-art performance in single-class, multi-class, and few-shot AD tasks across MVTec-AD, VisA, and Real-IAD, positioning it as a versatile and universal solution for AD. Remarkably, INP-Former also demonstrates some zero-shot AD capability. Furthermore, we propose a soft version of the INP Coherence Loss and enhance INP-Former by incorporating residual learning, leading to the development of INP-Former++. The proposed method significantly improves detection performance across single-class, multi-class, few-shot, and zero-shot settings, while demonstrating strong adaptability and superior performance in semi-supervised setting, further advancing the boundaries of universal anomaly detection.
\end{abstract}


}

\maketitle

\IEEEdisplaynontitleabstractindextext

%
\IEEEpeerreviewmaketitle

\IEEEraisesectionheading{
\section{Introduction}\label{sec:intro}}

\IEEEPARstart{U}{nsupervised} image anomaly detection (AD)~\cite{IM-IAD, ader, cao2024survey} seeks to identify abnormal patterns in images and localize anomalous regions by learning solely from normal samples. This technique has seen widespread application in industrial defect detection~\cite{MVTec-AD,VisA} and medical disease screening~\cite{MVFA}. Recently, various specialized tasks have emerged in response to real-world demands, from conventional single-class AD~\cite{Patchcore, luo2024template} to more advanced few-shot AD~\cite{regad,PSNet} and multi-class AD~\cite{uniad,guo2024dinomaly, PNPT}.

\begin{figure*}[t]
    \centering
    \includegraphics[width=0.98\linewidth]{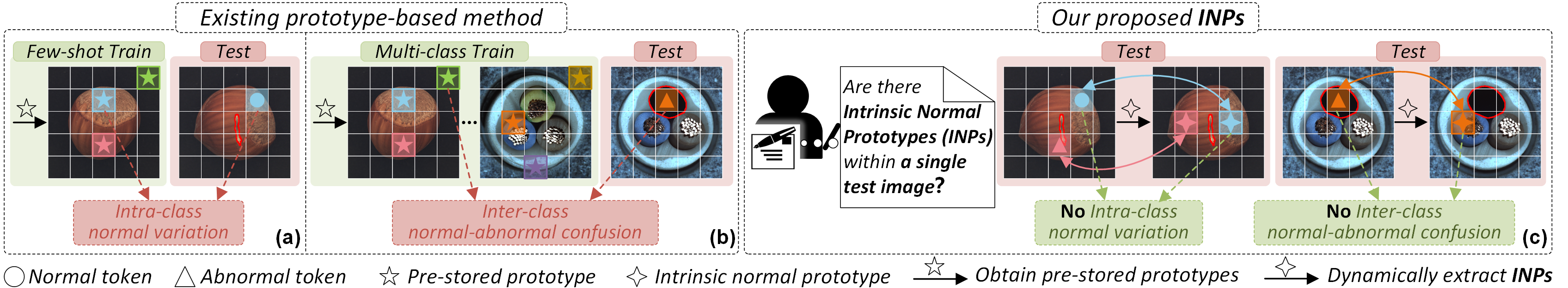}
    \vspace{-10.5pt}
    \caption{\textbf{Motivation for Intrinsic Normal Prototypes (\textit{INPs}).} (a) Pre-stored prototypes from few-shot normal samples may fail to represent all normal patterns. (b)  Pre-stored prototypes from one class can be similar to anomalies in another class. (c) The extracted \textbf{INPs} are concise yet well-aligned to the test image, alleviating the issues in (a) and (b).}
    \label{fig:teaser}
    \vspace{-3mm}
\end{figure*}

Although the composition of normal samples varies across these tasks, the fundamental principle remains unchanged: modeling normality in the training data and assessing whether a test image aligns with this learned normality. However, this approach can be limited due to \textit{misaligned normality} between the training data and the test image. For instance, prototype-based methods~\cite{Patchcore} extract representative normal prototypes to capture the normality of training samples. In few-shot AD, intra-class variance may lead to poorly aligned prototypes~\cite{regad}, \eg, hazelnuts in different appearances and positions, as shown in Fig.~\ref{fig:teaser}(a). Increasing the sample size can mitigate this problem but at the cost of additional prototypes and reduced inference efficiency. When there are multiple classes, \ie, multi-class AD, prototypes from one class may resemble anomalies from another, like the normal background of hazelnut is similar to the anomalies in cable in Fig.~\ref{fig:teaser}(b), leading to misclassification.

Several works have focused on extracting normality that is more aligned with the test image. For instance, some studies~\cite{regad,FOD,THFR} propose spatially aligning normality within a single class through geometrical transformations. However, spatial alignment is ineffective for certain objects, such as hazelnuts, which exhibit variations beyond spatial positions. Other approaches~\cite{PNPT,HVQ-Trans,HGAD} attempt to divide the normality in the training set into smaller, specific portions and then compare the test image to the corresponding portion of normality, but may still fail to find perfect alignment because of intra-class variances. 

Rather than attempting to extract more aligned normality from the training set, we propose addressing the issue of misaligned normality by leveraging the \textit{normality within the test image} itself as prototypes, which we term \textbf{Intrinsic Normal Prototypes (INPs)}. As illustrated in Fig.~\ref{fig:teaser}(c), normal patches within an anomalous test image can function as INPs, and anomalies can be easily detected by comparing them with these INPs. These INPs provide more concise and well-aligned prototypes to the anomalies than those learned from training data, as they typically share the same geometrical context and similar appearances with the abnormal regions within the testing image itself. Accordingly, we explore the prevalence of INPs in various AD scenarios and evaluate their potential to improve AD performance. While previous studies~\cite{INPtexture, yao2024global} have attempted to leverage INPs for anomaly detection by aggregating neighboring features to construct prototypes and enhance detection performance, these methods are generally constrained to zero-shot texture anomaly detection.

In this paper, we present a learnable INP-based framework, named \textbf{INP-Former}, for universal anomaly detection. First, we propose a learnable INP extractor that leverages a cross-attention mechanism to extract globally semantic normal features as INPs from a single image. To prevent the capture of anomalous regions during inference, we introduce an INP Coherence Loss, ensuring that the extracted INPs consistently and accurately represent the normal patterns in the test image. However, some weakly representative normal regions are challenging to model with a limited set of discrete INPs, resulting in background noise ({Fig.~\ref{fig:ablation_lsc}(c)}). To address this issue, we design an INP-guided Decoder that incorporates INPs into a reconstruction-based framework. This decoder aggregates the discrete INPs to ensure high-fidelity reconstruction of normal regions while effectively suppressing the reconstruction of anomalous regions, allowing the reconstruction error to act as a reliable anomaly score (second row in {Fig.~\ref{fig:anomalymap}}). Additionally, inspired by Focal Loss~\cite{focalloss} and Dinomaly~\cite{guo2024dinomaly}, we introduce a Soft Mining Loss that optimizes the reconstruction of challenging normal regions, \ie, hard samples, further improving reconstruction quality and anomaly detection performance.

\paragraph{Extension of the conference version~\cite{luo2025exploring}.}
To thoroughly validate the effectiveness of the proposed INP, we extend our conference version~\cite{luo2025exploring} to a semi-supervised AD~\cite{zhang2023PRN, yao2023explicit} setting, thereby further broadening the scope of universal anomaly detection.  First, we observe that the original INP coherence loss tends to converge to a shortcut solution, where most patch tokens are assigned to the same INP (Fig.~\ref{fig:lc_shortcut}(b)). To address this issue, we propose the Soft INP Coherence Loss, which ensures that the weighted combination of INPs aligns with each patch token, thereby extracting more representative INPs and further improving detection performance. Additionally, since the model is trained exclusively on normal samples, its ability to distinguish between normal and anomalous regions is limited, often leading to confusion between the two (Fig.~\ref{fig:ablation_rl}(b)). To mitigate this, we introduce a segmentation head, combined with pseudo-anomaly generation and a residual learning strategy, which amplifies the residual differences between normal and anomalous regions (Fig.~\ref{fig:ablation_rl}(c)). This improvement significantly enhances anomaly localization performance and extends our method to a semi-supervised anomaly detection setting. The extended framework, named \textbf{INP-Former++}, demonstrates exceptional detection performance across various settings, including few-shot, single-class, multi-class, and semi-supervised anomaly detection. Specifically, under the challenging multi-class setting of the widely used MVTec-AD~\cite{MVTec-AD} dataset, INP-Former++ achieves an image-level AUROC of \textbf{99.8}. In the few-shot setting (shot = 4), it still achieves an image-level AUROC of \textbf{97.9}, outperforming the existing Vision-Language-Model-based few-shot SOTA method, SimCLIP~\cite{deng2024simclip}, by \textcolor{purple}{1.5$\uparrow$}. In the semi-supervised setting (introducing 10 anomalous samples in addition to normal samples per category, also referred to as Semi10)~\cite{BiaS}, INP-Former++ achieves a remarkably high AUPRO of \textbf{96.8}. Beyond accuracy, we empirically find that images can be represented by only six INPs (Fig.~\ref{fig:prototypevis}), which greatly optimizes computational complexity. With more informative INPs, INP-Former++ also demonstrates improved zero-shot performance compared with the original INP-Former (Fig.~\ref{fig:anomalymapZS}(b) and (c)).
Moreover, Fig.~\ref{fig:scale} illustrates the strong scalability of INP-Former++, with performance steadily improving as more normal samples and a small number of real anomalies become available—\textbf{positioning INP-Former++ as a universal AD solution}. In summary, our main contributions are as follows:
\begin{itemize}
    \item To overcome the misalignment between prototypes extracted from training data and the normal patterns present in test images, we introduce an intuitive framework, INP-Former, which explores the normal patterns within a single test image as prototypes, termed INPs, for universal anomaly detection.
    \item We extend the original version of INP-Former~\cite{luo2025exploring} to a semi-supervised anomaly detection setting by integrating pseudo-anomaly generation and residual learning, thereby further expanding the scope of universal anomaly detection.
    \item To prevent shortcut solutions, where most features are assigned to the same INP, we extend INP Coherence Loss to Soft INP Coherence Loss. This approach aligns each feature with a weighted combination of INPs, facilitating the extraction of more representative INPs.
    \item To the best of our knowledge, the extended INP-Former++ framework is the first model to exhibit exceptional performance across single-class, few-shot, multi-class, and semi-supervised AD settings, while also demonstrating certain zero-shot detection capabilities, thereby laying a solid foundation for universal anomaly detection.
\end{itemize}

\begin{figure*}[t]
    \centering
    \includegraphics[width=0.99\linewidth]{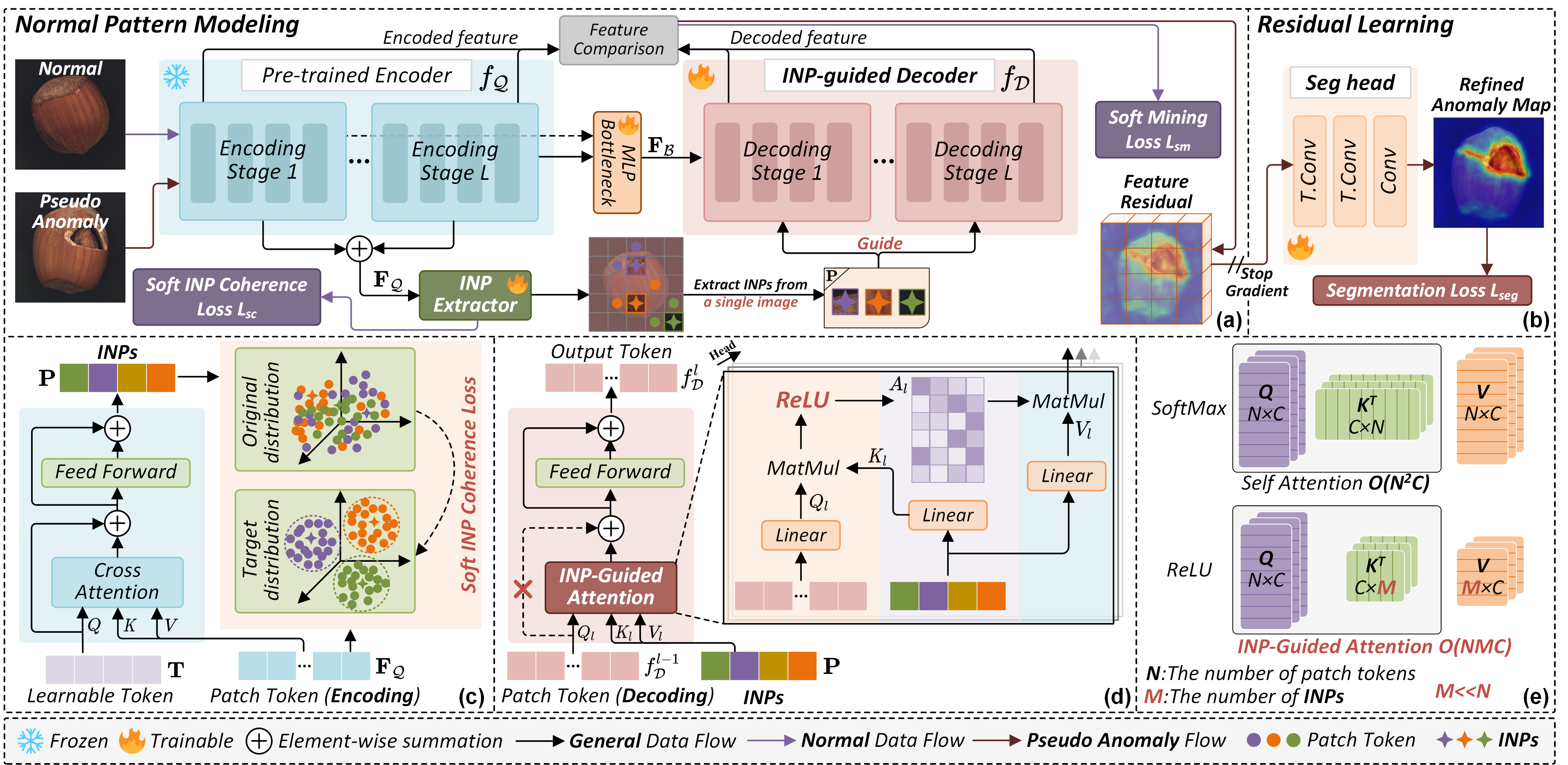}
    \vspace{-10pt}
    \caption{\textbf{Overview of our INP-Former++ framwork for universal anomaly detection.} Our model consists of a pre-trained Encoder, an INP Extractor, a Bottleneck, an INP-guided decoder, and a Segmentation (Seg) head. (a) \textbf{Normal Pattern Modeling}. The INP Extractor dynamically extracts intrinsic normal prototypes from a single image, which the INP-guided Decoder leverages to effectively suppress anomalous features. (b) \textbf{Residual Learning}. The pseudo anomalies are input into the pretrained encoder and INP-guided decoder to obtain feature residuals, which are then fed into the segmentation head to further enhance the residual differences between normal and anomalous regions.  (c) Detailed architecture of the INP Extractor. (d) Detailed architecture of each layer in the INP-guided Decoder. (e) Comparison of computational complexity between INP-guided Attention and Self Attention. It is important to note that the patch token \textbf{(Encoding)} and patch token \textbf{(Decoding)} refer to the patch tokens utilized during the encoding and decoding stages, respectively.}
    \label{fig:framework}
    \vspace{-3mm}
\end{figure*}
\section{Related Works}\label{sec:related_works}

\subsection{Universal Anomaly Detection}

There are numerous AD tasks, ranging from conventional single-class AD to recent few-shot, multi-class, and semi-supervised AD setups. We refer to these collectively as universal anomaly detection.
\subsubsection{Single-Class Anomaly Detection}
This setup was originally introduced by MVTec-AD~\cite{MVTec-AD} and involves developing distinct AD models for each class. Typically, images are embedded into a feature space using a pre-trained encoder, after which various schemes, such as reconstruction-based~\cite{AMI-Net,RealNet, RD4AD, RDplus}, prototype-based~\cite{Patchcore,FOD,jiang2022softpatch,wang2025softpatch+,hyun2024reconpatch}, and synthesizing-based~\cite{liu2023simplenet, zhang2023destseg, chen2024unified, PBAS, chen2025center} methods, are employed to learn the normality of the given class. Specifically, reconstruction-based methods employ self-supervised autoencoders to compress and restore input images or features, with convolutional neural networks, vision transformers~\cite{ViT}, and diffusion models~\cite{DDPM} commonly used as backbone architectures. Prototype-based methods model the distribution of normal data by storing pretrained features of normal samples in a memory bank. Synthesizing-based methods typically synthesize pseudo-anomalies by combining texture images with noise and subsequently utilize a discriminator network to predict the corresponding anomaly masks. While these approaches achieve strong performance, their reliance on class-specific models limits scalability when dealing with a wide range of classes.
\subsubsection{Few-shot Anomaly Detection}
In real-world applications, the availability of normal samples is often scarce, driving research towards few-shot AD techniques. Under such constraints, the limited normal data may fail to capture the full diversity of normal patterns. To mitigate this issue, methods leveraging spatial alignment~\cite{regad} or contrastive learning~\cite{PSNet} have been proposed to obtain more compact and discriminative representations of normality. More recently, Vision-Language Models (VLMs), such as CLIP~\cite{clip}, have emerged as powerful tools for few-shot AD tasks, owing to their broad pre-trained vision-language knowledge. These models can capture rich visual embeddings and facilitate text-image similarity matching, as demonstrated in methods like WinCLIP~\cite{WinClip}, AnomalyGPT~\cite{AnomalyGPT}, and InCTRL~\cite{InCTRL}. Additionally, certain methods, including AdaCLIP~\cite{AdaCLIP}, extend this capability to zero-shot AD settings by fully exploiting VLMs.
\subsubsection{Multi-Class Anomaly Detection}
Constructing individual models for each class is often resource-intensive, fueling interest in multi-class AD, also known as unified AD~\cite{uniad}, which aims to develop a single model capable of handling multiple categories. UniAD~\cite{uniad} introduce a pioneering unified reconstruction framework, followed by HVQ-Trans~\cite{HVQ-Trans}, which mitigated the shortcut learning issue via a vector quantization mechanism. Recent advancements, such as MambaAD~\cite{he2024mambaad} and Dinomaly~\cite{guo2024dinomaly}, have further boosted multi-class AD performance by integrating advanced architectures, including the State Space Model Mamba~\cite{mamba} and DINO~\cite{DINOV2-R}. However, these methods lack the ability to generate normal patterns that are specifically aligned with the test samples. In contrast, our approach extracts INPs directly from the test image, yielding well-aligned and precise normal representations for anomaly detection.
\subsubsection{Semi-Supervised Anomaly Detection}
In real-world scenarios, only a limited number of anomalous samples are typically available, motivating research on semi-supervised AD that aims to integrate the knowledge from scarce anomalies with abundant normal data. DeepSAD~\cite{deepsad} pioneered this direction by extending the unsupervised DeepSVDD~\cite{ruff2018deepsvdd} framework to a semi-supervised setting, leveraging a small set of anomalies during training. To further enhance performance, Ruff \textit{et al.}~\cite{ruff2020rethinking} refined the cross-entropy classification in DeepSAD to place greater emphasis on normal samples. To address the inherent bias from limited anomaly exposure, recent methodologies have introduced innovative architectural solutions. {Deviation-aware frameworks~\cite{pang2021explainable, zhang2020viral} employ margin-based optimization to enforce discriminative separation of latent representations between normal and anomalous classes. BiaS~\cite{BiaS} explicitly induces distributional divergence through teacher-student network pairs, while PRN~\cite{zhang2023PRN} proposes a multi-scale feature residual analysis to localize anomalous regions through hierarchical pattern deviations. Current approaches predominantly focus on joint optimization of normal pattern modeling and anomaly discrimination. Our work introduces a paradigm shift through \textit{decoupled learning}, where normal pattern modeling and residual-based anomaly amplification are optimized independently of each other. This structural decomposition addresses fundamental limitations in existing semi-supervised AD frameworks by preventing cross-phase interference and enabling targeted feature learning.}


\subsection{Prototype Learning}
Prototype learning~\cite{snell2017prototypical} aims to extract representative prototypes from a given training set, which are then used for classification by measuring their distances to a test sample in a metric space. This technique is widely used in few-shot learning~\cite{li2021adaptive}. Several AD methods also employ prototype learning. For example, PatchCore~\cite{Patchcore} extracts multiple normal prototypes to represent the normality of the training data, directly computing the minimal distances to the test sample for anomaly detection. Other approaches~\cite{park2020learning,lv2021learning,huang2022pixel,MemAE} incorporate prototypes into the reconstruction process to avoid the identical shortcut issue. Specifically, they replace the original inputs with combinations of learned normal prototypes, ensuring that the inputs to the reconstruction model contain only normal elements. However, these methods rely on pre-stored normal prototypes extracted from the training set, which can suffer from the misaligned normality problem. In contrast, our INPs are dynamically extracted from the test image, providing more aligned alternatives for normality representation.

\subsection{Residual Learning}
In recent years, residual learning has gained significant attention in the field of anomaly detection. PRN~\cite{zhang2023PRN} and OneNIP~\cite{onenip} utilize feature residuals to predict corresponding anomaly masks, facilitating precise anomaly localization. Additionally, some studies~\cite{pmad, yao2024resad} improve the model's performance in anomaly detection for unseen classes by leveraging residual features, as these residuals effectively eliminate class-specific attributes. {In this paper, we employ residual learning to amplify the discriminative boundary between normal and abnormal features, which not only enhances anomaly localization accuracy but also provides a feasible path for extending to semi-supervised AD}.
\section{Method: INP-Former++}\label{sec:method}
\subsection{Overview}

To fully exploit the advantages of INPs in anomaly detection, we introduce INP-Former++, as depicted in Figs.~\ref{fig:framework}(a) and (b). 
{The framework employs a decoupled training paradigm consisting of \textit{Normal Pattern Modeling} and \textit{Residual Learning}, where the optimization of these two components is independent of each other. The former relies solely on normal data to simultaneously learn INP extraction and INP-guided reconstruction, deliberately avoiding the use of pseudo anomalies to prevent potential bias. In contrast, the latter utilizes pseudo anomalies to further enhance the discriminative boundary between normal and anomalous features.} {Following RD4AD~\cite{RD4AD} and Dinomaly~\cite{guo2024dinomaly}, we adopt a feature reconstruction framework.} Specifically, INP-Former++ comprises five key modules: a fixed pre-trained Encoder $\mathcal{Q}$, an INP Extractor $\mathcal{E}$, a Bottleneck $\mathcal{B}$, an INP-guided Decoder $\mathcal{D}$, and a Segmentation (Seg) Head $\mathcal{H}$. The input image $\mathbf{I}\in\mathbb{R}^{H\times W\times 3}$ is first processed by the pre-trained Encoder $\mathcal{Q}$ to extract multi-scale latent features $f_{\mathcal{Q}}=\{f_{\mathcal{Q}}^{1}, \dots, f_{\mathcal{Q}}^{L}|f_{\mathcal{Q}}^{l}\in\mathbb{R}^{N\times C}, N=\frac{HW}{k^2}\}$, where $k$ represents the downsampling factor. Next, the INP Extractor $\mathcal{E}$ extracts $M$ INPs $\mathbf{P}=\{p_{1},\dots, p_{M}|p_{m}\in\mathbb{R}^{C}\}$ from the pre-trained features, with an INP coherence loss ensuring that the extracted INPs consistently represent normal features during testing. The Bottleneck $\mathcal{B}$ subsequently fuses the multi-scale latent features, producing the fused output $F_{\mathcal{B}}=\mathcal{B}(f_{\mathcal{Q}})$. Following the bottleneck, the extracted INPs are utilized to guide the Decoder $\mathcal{D}$ to yield reconstruction outputs $f_{\mathcal{D}}=\{f_{\mathcal{D}}^{1}, \dots, f_{\mathcal{D}}^{L}|f_{\mathcal{D}}^{l}\in\mathbb{R}^{N\times C}\}$ 
with only normal patterns, thus the reconstruction error between $f_{\mathcal{Q}}$ and $f_{\mathcal{D}}$ can serve as the feature residual. It is worth noting that we adopt the group-to-group feature reconstruction strategy introduced in Dinomaly~\cite{guo2024dinomaly}. Subsequently, given the pseudo or real anomalous image $\mathbf{I}^{a}$, we follow a similar procedure to obtain the corresponding feature residual $f_{res}^{a}$. This residual is then passed through the Seg Head $\mathcal{H}$ to further enhance the residual contrast between normal and abnormal regions, yielding a more refined anomaly map.  


\subsection{INP Extractor}
Existing prototype-based methods~\cite{Patchcore, MemAE, MNAD} store local normal features from the training data and compare them with test images. However, the misaligned normality between these pre-stored prototypes and the test images, and the lack of global information lead to suboptimal detection performance. To address these limitations, we propose the INP Extractor to dynamically extract INPs with global information from the test image itself. 

Specifically, as illustrated in Fig.~\ref{fig:framework}(c), instead of extracting representative local features as in PatchCore~\cite{Patchcore}, we employ cross attention to aggregate the global semantic information of the pre-trained features $\mathbf{F}_{\mathcal{Q}}\in\mathbb{R}^{N\times C}$ with $M$ learnable tokens $\mathbf{T}=\{t_{1},\dots, t_{M}|t_{m}\in\mathbb{R}^{C}\}$. Here $\mathbf{F}_\mathcal{Q}$ is used as the key-value pairs, while $\mathbf{T}$ serve as the query, allowing $\mathbf{T}$ to linearly aggregate $\mathbf{F}_\mathcal{Q}$ into INPs $\mathbf{P}=\{p_{1},\dots, p_{M}|p_{m}\in\mathbb{R}^{C}\}$. 
\begin{equation}
\begin{aligned}
    &\mathbf{F}_{\mathcal{Q}} = \operatorname{sum}(\{f_{\mathcal{Q}}^{1},\dots,f_{\mathcal{Q}}^{L}\})\\
    &Q = \mathbf{T}W^{Q}, K = \mathbf{F}_{\mathcal{Q}}W^{K}, V=\mathbf{F}_{\mathcal{Q}}W^{V}\\
    &\mathbf{{T}'} =\operatorname{Attention}(Q,K,V) + \mathbf{T}\\ 
    &\mathbf{P} = \operatorname{FFN}(\mathbf{{T}'}) + \mathbf{{T}'}
\end{aligned}
\end{equation}
\noindent where $\operatorname{sum}(\cdot)$ denotes the element-wise summation. $Q\in\mathbb{R}^{M\times C}$ and $K,V\in\mathbb{R}^{N\times C}$ represent the query, key and value, respectively. $W^{Q},W^{K},W^{V}\in\mathbb{R}^{C\times C}$ are the learnable projection parameters for $Q,K,V$. $\operatorname{FFN}(\cdot)$ represents the feed-forward network.

\noindent\textbf{INP Coherence Loss.} To ensure that INPs coherently represent normal features while minimizing the capture of anomalous features during the testing process, we propose an INP coherence loss $\mathcal{L}_{c}$  to minimize the distances between individual normal features and the corresponding nearest INP.
\begin{equation}
\label{eq:lc}
\begin{aligned}
    &d_{i} = \underset{m\in\{1,\dots,M\}}{\operatorname{min}}(1-\mathcal{S}(\mathbf{F}_{\mathcal{Q}}(i),p_{m}))\\
    &\mathcal{L}_{c} = \frac{1}{N}\sum_{i=1}^{N}d_{i}
\end{aligned}
\end{equation}
\noindent where $\mathcal{S}(\cdot, \cdot)$ denotes the cosine similarity. $d_i$ represents the distance between the query feature $\mathbf{F}_{\mathcal{Q}}(i)$ and the corresponding nearest INP item. Fig.~\ref{fig:ablation_lsc}(c) visually illustrates the effectiveness of $\mathcal{L}_{c}$.\\
\noindent\textbf{Soft INP Coherence Loss.} However, we observe that the original INP coherence loss tends to fall into a shortcut solution, where all patch tokens are assigned to the same INP, as illustrated in {Fig.~\ref{fig:lc_shortcut}(b)}. To address this issue, we extend it to a soft INP coherence loss $\mathcal{L}_{sc}$, which encourages a weighted combination of INPs to align with each patch token. Instead of regional cosine similarity, we adopt global cosine similarity~\cite{guo2023recontrast} to better capture the overall structure of the feature during INP extraction.
\begin{equation}
    \begin{aligned}
    &w_{i,m}=\frac{ \operatorname{exp}(\mathcal{S}(\mathcal{\mathbf{F}}_{\mathcal{Q}}(i),p_m))}{\sum_{j=1}^{M} \operatorname{exp}(\mathcal{S}(\mathcal{\mathbf{F}}_{\mathcal{Q}}(i),p_j))}\\
    &\hat{\mathbf{F}}_\mathcal{Q}(i) = \sum_{m=1}^{M}w_{i,m}p_{m}\\
    & \mathcal{L}_{sc} = 1-\mathcal{S}(vec(\mathbf{F}_{\mathcal{Q}}), vec(\hat{\mathbf{F}}_{\mathcal{Q}}))
\end{aligned}
\end{equation}
where $w_{i,m}$ denotes the normalized weight of the $m$-th INP for the $i$-th query feature $\mathbf{F}_{\mathcal{Q}}(i)$, and $\hat{\mathbf{F}}_{\mathcal{Q}}(i)$  is the reconstructed patch token derived from the weighted combination of INPs. The operation $vec(\cdot)$ refers to the flattening operation. {Fig.~\ref{fig:lc_shortcut}(c)} illustrates that $\mathcal{L}_{sc}$ does not fall into the shortcut solution mentioned above. Moreover, the comparison between {Fig.~\ref{fig:ablation_lsc}(c) and (d)} further demonstrates the superiority of $\mathcal{L}_{sc}$ over $\mathcal{L}_{c}$.
\begin{table}[!t]
\centering
\caption{Comparison of \textbf{computational cost} and \textbf{memory usage}.}
\label{table:comparison of computational}
\vspace{-10pt}
\fontsize{11}{14}\selectfont{
\resizebox{1.0\linewidth}{!}{
\begin{tabular}{c|cc}
\toprule[1.5pt]
           & \multicolumn{2}{c}{Number of multiplicaiton and addition}          \\ \midrule
Calculation & \multicolumn{1}{c|}{Vanilla Self Attention} & \textbf{INP-guided Attention} \\ \midrule
$A_{l}=Q_{l}(K_{l})^{T}$      & \multicolumn{1}{c|}{943 496 960}            & \textbf{7 220 640}            \\ \midrule
${f_{\mathcal{D}}^{l-1}}'=A_{l}V_{l}$        & \multicolumn{1}{c|}{943 509 504}            & \textbf{6 623 232}            \\ \midrule
 & \multicolumn{2}{c}{Memory usage (MB)}                                         \\ \midrule
$Q_l$/$K_l$/$V_l$/$A_l$    & \multicolumn{1}{c|}{2.30/2.30/2.30/2.34}                & 2.30/\textbf{0.018}/\textbf{0.018}/\textbf{0.018}              \\ \bottomrule[1.5pt]
\end{tabular}}}
\vspace{-3mm}
\end{table}

\subsection{INP-guided Decoder}
While we can use the distance between testing features and their corresponding weighted combination of INPs for anomaly detection, as illustrated in {Fig.~\ref{fig:ablation_lsc}(d)}, certain low-representative normal regions are difficult to model using a simple linear combination of INPs, leading to noisy and coarse distance maps between these INPs and testing features. To address this issue, we propose the INP-guided Decoder, aiming to reconstruct these low-representative normal regions through a learnable and complex nonlinear combination of multiple discrete INPs and suppress the reconstruction of anomalous regions. Additionally, this decoder provides a token-wise discrepancy that can be directly leveraged for anomaly detection. As shown in Fig.~\ref{fig:framework}(d), INPs are incorporated into this decoder to guide the reconstruction process. Since INPs exclusively represent normal patterns in test images, we employ the extracted INPs as key-value pairs, ensuring that the output is a linear combination of normal INPs, thereby effectively suppressing the reconstruction of anomalous queries, \ie, the idenfical mapping issue~\cite{uniad}. Furthermore, we find that the first residual connection can directly introduce anomalous features into the subsequent reconstruction, so we remove this connection in our INP-guided Decoder. Following the previous work~\cite{huang2024sparse}, we also employ the ReLU activation function to mitigate the influence of weak correlations and noise on the attention maps.

Mathematically, let $f_{\mathcal{D}}^{l-1}\in\mathbb{R}^{N\times C}$ denotes the output latent features from previous decoding layer. The output $f_{\mathcal{D}}^{l}\in\mathbb{R}^{N\times C}$ of the $l_{th}$ decoding layer is formulated as:
\begin{equation}
    \begin{aligned}
    &Q_{l} = f_{\mathcal{D}}^{l-1}W^{Q}_{l}, K_{l} = \mathbf{P}W^{K}_{l}, V_{l}=\mathbf{P}W^{V}_{l}\\
    &{f_{\mathcal{D}}^{l-1}}'=A_{l}V_{l}, A_{l} = \operatorname{ReLU}(Q_{l}(K_{l})^T)\\
    &f_{\mathcal{D}}^{l} = \operatorname{FFN}({f_{\mathcal{D}}^{l-1}}') + {f_{\mathcal{D}}^{l-1}}'
\end{aligned}
\end{equation}
where $Q_l\in\mathbb{R}^{N\times C}$ and $K_l, V_l\in\mathbb{R}^{M\times C}$ denote the query, key and value of the $l_{th}$ decoding layer. $W_{l}^{Q},W_{l}^{K},W_{l}^{V}\in\mathbb{R}^{C\times C}$ denote the learnable projection parameters for $Q_{l},K_{l},V_{l}$. $A_{l}\in\mathbb{R}^{N\times M}$ represent the attention map.

\noindent\textbf{Attention Complexity Analysis}: As depicted in Fig.~\ref{fig:framework}(e), the computational complexity of vanilla self-attention is $\mathcal{O}(N^2C)$, while its memory usage is $\mathcal{O}(N^2)$. In contrast, our INP-guided Attention reduces both the computational complexity and memory usage to $\mathcal{O}(NMC)$ and $\mathcal{O}(NM)$, respectively, which can be approximated as $\mathcal{O}(NC)$ and $\mathcal{O}(N)$ due to $M\ll N$. Tab.~\ref{table:comparison of computational} presents a detailed comparison of vanilla self-attention and INP-guided Attention in terms of computational cost and memory usage, with INP-guided Attention reducing both by nearly 99.2\%. Sec.~\ref{sec-complex} provides a comparison of the overall complexity between INP-Former++ and other methods. The lightweight variant of INP-Former++ demonstrates comparable efficiency to MambaAD~\cite{he2024mambaad}, while delivering better performance.


\subsection{Soft Mining Loss}

Inspired by Focal Loss~\cite{focalloss} and Dinomaly~\cite{guo2024dinomaly}, different regions should be assigned varying weights based on their optimization difficulty. Accordingly, we propose Soft Mining Loss to encourage the model to focus more on difficult regions.

Intuitively, the ratio of the reconstruction error of an individual normal region to the average reconstruction error of all normal regions can serve as an indicator of optimization difficulty. To comprehensively assess both the direction and magnitude of feature vectors, we employ a combination of cosine similarity and mean squared error (MSE) as dual criteria for the reconstruction loss, with MSE providing a solid foundation for the subsequent residual learning. Moreover, following Dinomaly~\cite{guo2024dinomaly} and ReContrast~\cite{guo2023recontrast}, we modify the feature gradients instead of applying reweighting strategies~\cite{CDO}, aiming to preserve the global structure of the feature point manifolds. Specifically, given the encoder $f_{\mathcal{Q}}^{l}\in \mathbb{R}^{h \times w \times C}$ and decoder $f_{\mathcal{D}}^{l}\in \mathbb{R}^{h \times w \times C}$ features at layer $l$, where $h=\frac{H}{k}$ and $w=\frac{W}{k}$, let $M^{l}_{cos}$, $M^{l}_{mse}\in \mathbb{R}^{h \times w}$ denote the regional cosine distance~\cite{guo2023recontrast} and MSE. Our soft mining loss $\mathcal{L}_{sm}$ is defined as follows:{
\begin{equation}
\begin{aligned}
        &\mathbf{w}_{t}^{l}(i,j)=\left[\frac{M_{t}^{l}(i,j)}{u(M_{t}^{l})}\right]^\gamma,t\in\{cos,mse\} \\
           &\nabla\hat{f}_{\mathcal{D}}^{l,t}(i,j)=\nabla{f}_{\mathcal{D}}^{l}(i,j)\cdot{\mathbf{w}_{t}^{l}(i,j)},t\in\{cos,mse\}\\
    &\mathcal{L}_{sm}^{cos}=\frac{1}{L}\sum_{l=1}^{L}\left(1-\mathcal{S}(vec(f_{\mathcal{Q}}^{l}), vec(\hat{f}_{\mathcal{D}}^{l, cos}))\right)\\
    &\mathcal{L}_{sm}^{mse}=\frac{1}{Lhw}\sum_{l=1}^{L}\sum_{i=1}^{h}\sum_{j=1}^{w}\left \|f_\mathcal{Q}^{l}(i,j)-\hat{f}_{\mathcal{D}}^{l, mse}(i,j)  \right \| _{2}^{2}\\
      & \mathcal{L}_{sm}=\mathcal{L}_{sm}^{cos} + \mathcal{L}_{sm}^{mse}
\end{aligned}
\end{equation}}where $u(M_{t}^{l})$ represents the average regional cosine distance or MSE within a batch, $\gamma\geq0$ denotes the temperature hyperparameter,  {$\nabla$ denotes the gradient operator}, and $\left\|\cdot\right\|_{2}$ denotes the $\ell_2$ norm. The overall training loss of our INP-Former++ for normal pattern modeling can be expressed as follows:{
\begin{equation}
\mathcal{L}_{npm}=\mathcal{L}_{sm}+\lambda\mathcal{L}_{sc}
\end{equation}}where $\lambda$ denotes the weight that balances the relative importance of the two losses.


\subsection{Residual Learning}
Training exclusively on normal data can lead to limited discrimination between normal and anomalous regions.
To mitigate this issue, as illustrated in Fig.~\ref{fig:framework}(b), we draw inspiration from OneNIP~\cite{onenip} and introduce a segmentation head, combining pseudo-anomaly generation and residual learning strategies to amplify the residual differences between normal and anomalous regions. {It is important to note that, during residual learning, we stop the gradient of feature residual, allowing only the segmentation head $\mathcal{H}$ to be optimized.} This strategy intends to prevent the reconstruction model from being biased by the pseudo-anomalies. Specifically, similar to DRAEM~\cite{zavrtanik2021draem}, we utilize Perlin noise~\cite{perlin1985image} and texture images from DTD~\cite{DTDDATASET} dataset to synthesize out-of-distribution pseudo-anomalies $\mathbf{I}^{a}\in\mathbb{R}^{H\times W\times 3}$ and the corresponding pixel-level mask $\mathbf{M}^{a}\in\mathbb{R}^{H\times W}$. Let $f_{\mathcal{Q}}^{l,a}\in \mathbb{R}^{h \times w \times C}$ and $f_{\mathcal{D}}^{l,a}\in \mathbb{R}^{h \times w \times C}$ represent the corresponding encoder and decoder feature at layer $l$, and their feature residuals $f_{res}^{a}\in \mathbb{R}^{h \times w \times C}$ can be computed as follows:
\begin{equation}
\begin{aligned}
    \begin{aligned}
f_{res}^a(i,j) = = \frac{1}{L} \sum_{l=1}^{L} &\left[ \left(1 - \mathcal{S}(f_{\mathcal{Q}}^{l,a}(i,j), f_{\mathcal{D}}^{l,a}(i,j))\right) \right. \\
& \left. \odot \left| f_{\mathcal{Q}}^{l,a}(i,j) - f_{\mathcal{D}}^{l,a}(i,j) \right| \right]
\end{aligned}
\end{aligned}
\end{equation}
where $\odot$ denotes the element-wise product, and $\left|\cdot\right|$ represents the absolute value operation. Subsequently, $f_{res}^{a}$ is fed into the segmentation head, followed by upsampling to the original image size using bilinear interpolation, ultimately yielding the predicted anomaly mask $\mathbf{M}^{a}_{pred}\in\mathbb{R}^{H\times W}$. Considering the extreme imbalance in the number of normal and anomalous pixels, we adopt Dice loss~\cite{diceloss} as the segmentation loss function $\mathcal{L}_{seg}$:\begin{equation}
    \mathcal{L}_{seg} = 1 - \frac{2 \cdot \sum_{i=1}^{H} \sum_{j=1}^{W} \mathbf{M}^{a}_{pred}({i,j}) \cdot \mathbf{M}^{a}(i,j)}{\sum_{i=1}^{H} \sum_{j=1}^{W} \mathbf{M}^{a}_{pred}({i,j})^2 + \mathbf{M}^{a}(i,j)^2}
\end{equation}
\indent The incorporation of residual learning further extends the applicability of our method to \textbf{semi-supervised anomaly detection}. A straightforward yet effective approach to enhance anomaly diversity is to integrate a small amount of available real anomalies into the pseudo ones. In this study, following~\cite{zhang2023PRN, yao2023explicit}, we augment specific anomalous regions from these real anomalies and randomly embed them into normal images, thereby further enriching the diversity of the anomaly data.\\
\indent During the testing phase, we combine the reconstruction error $\mathbf{A}_{rec}\in\mathbb{R}^{h\times w}$ with the anomaly mask predicted by the segmentation head to obtain the final anomaly map $\mathbf{A}\in\mathbb{R}^{H\times W}$.
\begin{equation}
\begin{aligned}
    \mathbf{A}_{rec}(i,j) = \frac{1}{L}\sum_{l=1}^{L} &\frac{1}{2} \cdot \left[ \left(1 - \mathcal{S}(f_{\mathcal{Q}}^{l,a}(i,j), f_{\mathcal{D}}^{l,a}(i,j))\right) \right. \\
    & \quad \left. + \left\| f_{\mathcal{Q}}^{l,a}(i,j) - f_{\mathcal{D}}^{l,a}(i,j) \right\|_2 \right] \\
    \mathbf{A} &= \frac{\Theta(\mathbf{A}_{rec}) + \mathbf{M}_{pred}^{a}}{2}
\end{aligned}
\end{equation}
where $\Theta(\cdot)$ denotes bilinear upsampling to the original image resolution. Following~\cite{MSFLOW}, we obtain the image-level anomaly score by averaging the top 1\% values in the final anomaly map $\mathbf{A}$.\\


\begin{table}[h]
\centering
\caption{Comprehensive details of the datasets.}
\label{table:dataset}
\vspace{-10pt}
\fontsize{10.5}{14}\selectfont{
\resizebox{0.95\linewidth}{!}{
\begin{tabular}{c|c|c|cc}
\toprule[1.5pt]
\multirow{2}{*}{Dataset} & Category & Train  & \multicolumn{2}{c}{Test} \\ \cline{3-5} 
                         & Number   & Normal & Anomaly     & Normal     \\ \midrule
MVTec-AD~\cite{MVTec-AD}                 & 15       & 3,629   & 1,258        & 467        \\
VisA~\cite{VisA}                     & 12       & 8,659   & 962         & 1,200       \\
Real-IAD~\cite{real-iad}                 & 30       & 36,345  & 51,329       & 63,256      \\ \midrule
Uni-Medical~\cite{zhang2023exploring}              & 3        & 13,339  & 4,499        & 2,514       \\ \bottomrule[1.5pt]
\end{tabular}}}
\end{table}

\section{Experiments}\label{sec:experiments}
\subsection{Experimental Settings}
\subsubsection{Datasets}
We conduct a comprehensive analysis of the proposed INP-Former++ on three widely used AD industrial datasets——\textbf{MVTec-AD}~\cite{MVTec-AD}, \textbf{VisA}~\cite{VisA}, and \textbf{Real-IAD}~\cite{real-iad}——as well as a medical AD dataset, \textbf{Uni-Medical}~\cite{zhang2023exploring}. \textbf{MVTec-AD} comprises 15 categories, with 3,629 normal images for training and 1,982 anomalous images along with 498 normal images for testing.  \textbf{VisA} includes 12 object categories, containing 8,659 normal images for training and 962 normal images along with 1,200 anomalous images for testing. \textbf{Real-IAD} consists of 30 different objects, with 36,645 normal images for training and 63,256 normal images along with 51,329 anomalous images for testing. \textbf{Uni-Medical} contains 13,339 training images and 7,013 test images (4,499 anomalous and 2,514 normal) across three object types: brain CT, liver CT, and retinal OCT. The detailed information of the datasets is summarized in Tab.~\ref{table:dataset}.
\subsubsection{Metrics}
Following existing works~\cite{he2024mambaad, guo2024dinomaly}, we use the Area Under the Receiver Operating Characteristic Curve (AUROC), Average Precision (AP), and F1-score-max (F1\_max) to evaluate anomaly detection and localization. For anomaly localization specifically, we use Area Under the Per-Region-Overlap (AUPRO) as an additional metric.
\subsubsection{Implementation Details}
INP-Former++ adopts ViT-Base/14 with DINO2-R~\cite{DINOV2-R} weights as the default pre-trained encoder. Building on {Dinomaly~\cite{guo2024dinomaly}}, we employ a group-to-group supervision approach by aggregating the features from selected layers to form distinct groups. In our study, we define two groups: the features from layers 3 to 6 of ViT-Base constitute one group, while those from layers 7 to 10 form the other. The number $M$ of INPs is set to six by default. The INP Extractor includes a standard Vision Transformer block. The layer number of INP-guided decoder is eight. The segmentation head consists of two transposed convolution layers and a convolution layer. All input images are resized to $448^2$ and then center-cropped to $392^2$. The hyperparameters $\gamma$ and $\lambda$ are set to 3.0 and 0.2, respectively. We utilize the StableAdamW~\cite{wortsman2023stable} optimizer with a learning rate $5e^{-4}$ and a weight decay of $1e^{-4}$ for 200 epochs. Notably, the above hyperparameters do not require any adjustment across the three datasets. In the few-shot setting, we apply data augmentation techniques similar to those used in RegAD~\cite{regad}. The experimental code is implemented in Python 3.8 and PyTorch 2.0.0 (CUDA 11.8) and runs on an NVIDIA GeForce RTX 4090 GPU (24GB).

\subsection{Main Results}
\begin{table*}[!ht]
\centering
\caption{\textbf{Multi-class} anomaly detection performance on different AD datasets. The best in \textbf{bold}, the
second-highest is \underline{underlined}.}
\label{table:multi-class-main-performance}
\vspace{-10pt}
\fontsize{10}{14}\selectfont{
\resizebox{\textwidth}{!}
{\begin{tabular}{c|cccccc}
\toprule[1.5pt]
Dataset~$\rightarrow$    & \multicolumn{2}{c|}{MVTec-AD~\cite{MVTec-AD}}                                                     & \multicolumn{2}{c|}{VisA~\cite{VisA}}                                                         & \multicolumn{2}{c}{Real-IAD~\cite{real-iad}}         \\ \midrule
Metric~$\rightarrow$     & \multicolumn{6}{c}{Image-level(I-AUROC/I-AP/I-F1\_max)\hspace{10mm}Pixel-level(P-AUROC/P-AP/P-F1\_max/AUPRO)}                                                                                                                        \\  \cmidrule{2-7}
Method~$\downarrow$     & Image-level    & \multicolumn{1}{c|}{Pixel-level}                                 & Image-level    & \multicolumn{1}{c|}{Pixel-level}                                 & Image-level    & Pixel-level         \\ \midrule
RD4AD~\cite{RD4AD}      & 94.6/96.5/95.2 & \multicolumn{1}{c|}{96.1/48.6/53.8/91.1}                         & 92.4/92.4/89.6 & \multicolumn{1}{c|}{98.1/38.0/42.6/91.8}                         & 82.4/79.0/73.9 & 97.3/25.0/32.7/89.6 \\
UniAD~\cite{uniad}      & 96.5/98.8/96.2 & \multicolumn{1}{c|}{96.8/43.4/49.5/90.7}                         & 88.8/90.8/85.8 & \multicolumn{1}{c|}{98.3/33.7/39.0/85.5}                         & 83.0/80.9/74.3 & 97.3/21.1/29.2/86.7 \\
SimpleNet~\cite{liu2023simplenet}   & 95.3/98.4/95.8 & \multicolumn{1}{c|}{96.9/45.9/49.7/86.5}                         & 87.2/87.0/81.8 & \multicolumn{1}{c|}{96.8/34.7/37.8/81.4}                         & 57.2/53.4/61.5 & 75.7/2.8/6.5/39.0   \\
DeSTSeg~\cite{zhang2023destseg}    & 89.2/95.5/91.6 & \multicolumn{1}{c|}{93.1/54.3/50.9/64.8}                         & 88.9/89.0/85.2 & \multicolumn{1}{c|}{96.1/39.6/43.4/67.4}                         & 82.3/79.2/73.2 & 94.6/37.9/41.7/40.6 \\
DiAD~\cite{diad}      & 97.2/99.0/96.5 & \multicolumn{1}{c|}{96.8/52.6/55.5/90.7}                         & 86.8/88.3/85.1 & \multicolumn{1}{c|}{96.0/26.1/33.0/75.2}                         & 75.6/66.4/69.9 & 88.0/2.9/7.1/58.1   \\
MambaAD~\cite{he2024mambaad}    & 98.6/99.6/97.8 & \multicolumn{1}{c|}{97.7/56.3/59.2/93.1}                         & 94.3/94.5/89.4 & \multicolumn{1}{c|}{98.5/39.4/44.0/91.0}                         & 86.3/84.6/77.0 & 98.5/33.0/38.7/90.5 \\
Dinomaly\footref{reproduce}~\cite{guo2024dinomaly}   & 99.6/\underline{99.8}/99.1 & \multicolumn{1}{c|}{98.3/65.6/66.9/94.4}                         & \underline{98.7}/\underline{98.8}/95.9 & \multicolumn{1}{c|}{98.7/\underline{48.6}/\underline{52.3}/\underline{95.1}}                         & 89.4/86.8/80.2 & 99.0/39.8/44.9/93.9 \\ \midrule
\textbf{INP-Former}\footref{reproduce}~\cite{luo2025exploring} & \underline{99.7}/\textbf{99.9}/\underline{99.2} & \multicolumn{1}{c|}{\underline{98.5}/\underline{68.0}/\underline{67.7}/\underline{94.7}} & \textbf{98.9}/\textbf{99.0}/\textbf{96.6} & \multicolumn{1}{c|}{\underline{99.0}/46.8/51.1/95.0} & \underline{90.5}/\underline{88.1}/\underline{81.5} & \underline{99.1}/\underline{44.2}/\underline{47.9}/\underline{94.9} \\
\rowcolor{Light}
\textbf{INP-Former++} & \textbf{99.8}/\textbf{99.9}/\textbf{99.3} & \multicolumn{1}{c|}{\cellcolor{Light}\textbf{98.7}/\textbf{75.6}/\textbf{71.6}/\textbf{96.0}} & \textbf{98.9}/\textbf{99.0}/\underline{96.4} & \multicolumn{1}{c|}{\cellcolor{Light}\textbf{99.1}/\textbf{52.7}/\textbf{55.2}/\textbf{96.0}} & \textbf{90.7}/\textbf{88.3}/\textbf{81.6} & \textbf{99.2}/\textbf{50.2}/\textbf{50.9}/\textbf{95.1} \\ \bottomrule[1.5pt]
\end{tabular}}}
\vspace{-3mm}
\end{table*}

\begin{figure*}[!ht]
    \centering
    \includegraphics[width=\linewidth]{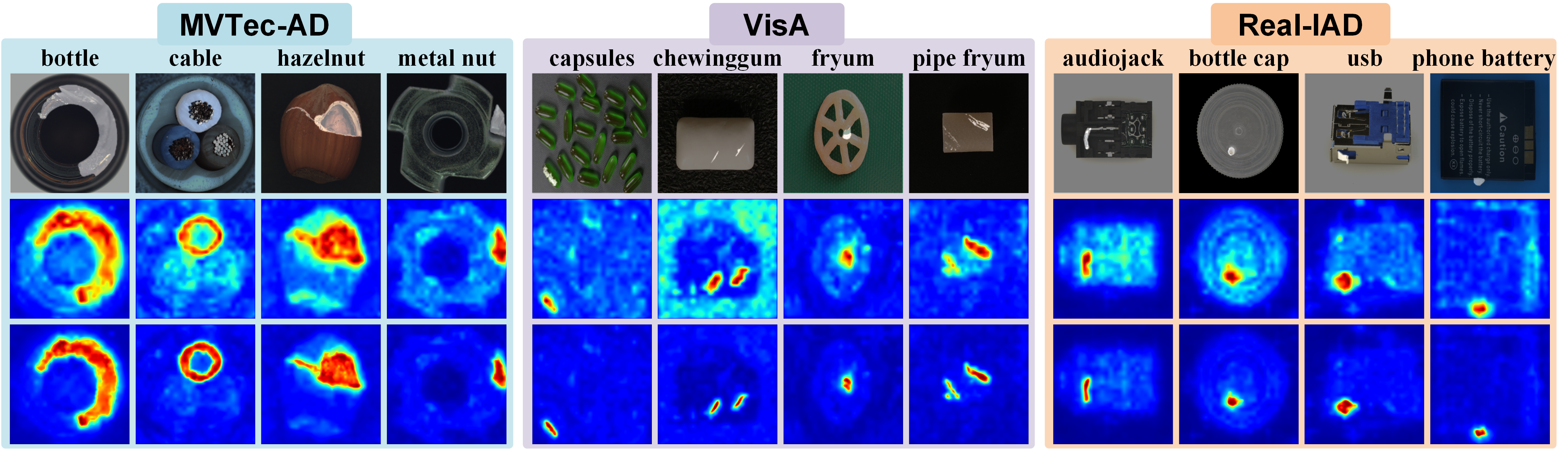}
    \vspace{-20pt}
    \caption{\textbf{Qualitative results of anomaly localization} on the MVTec-AD~\cite{MVTec-AD}, VisA~\cite{VisA}, and Real-IAD~\cite{real-iad} datasets for \textbf{multi-class anomaly detection}. From top to bottom, we show the input anomalous images along with their ground truth masks, the predicted anomaly maps from INP-Former, and the results from the improved \textbf{INP-Former++} model.}
    \label{fig:anomalymap}
\end{figure*}
\subsubsection{Multi-Class Anomaly Detection}
We compare INP-Former++ with state-of-the-art (SOTA) methods in the multi-class setting, including reconstruction-based approaches such as RD4AD~\cite{RD4AD}, UniAD~\cite{uniad}, DiAD~\cite{diad}, MambaAD~\cite{he2024mambaad}, and Dinomaly\footnote{\label{reproduce}Due to the \href{https://github.com/guojiajeremy/Dinomaly/issues/14}{issue} with the pixel-level metrics presented in the original paper, we reproduce the results using the publicly available source code.}~\cite{guo2024dinomaly}, as well as synthesizing-based methods like SimpleNet~\cite{liu2023simplenet} and DeSTSeg~\cite{zhang2023destseg}. Furthermore, we include our original conference version, INP-Former\footref{reproduce}~\cite{luo2025exploring}, for comprehensive evaluation.\\
\indent Tab.~\ref{table:multi-class-main-performance} presents a detailed quantitative comparison, demonstrating that INP-Former++ consistently achieves SOTA performance across three industrial AD datasets. Specifically, on the widely used MVTec-AD, it achieves I-AUROC/I-AP/I-F1\_{max} scores of \textbf{99.8}/\textbf{99.9}/\textbf{99.3}  and P-AUROC/P-AP/P-F1\_max/AUPRO scores of \textbf{98.7}/\textbf{75.6}/\textbf{71.6}/\textbf{96.0}, surpassing the second-best method, Dinomaly, by \textcolor{purple}{0.2$\uparrow$}/\textcolor{purple}{0.1$\uparrow$}/\textcolor{purple}{0.2$\uparrow$} in image-level scores and \textcolor{purple}{0.4$\uparrow$}/\textcolor{purple}{10.0$\uparrow$}/\textcolor{purple}{4.7$\uparrow$}/\textcolor{purple}{1.6$\uparrow$} in pixel-level scores. On the VisA dataset, the model attains image-level scores of \textbf{98.9}/\textbf{99.0}/\underline{96.4} and pixel-level scores of \textbf{99.1}/\textbf{52.7}/\textbf{55.2}/\textbf{96.0}, outperforming the second-best method by \textcolor{purple}{0.2$\uparrow$}/\textcolor{purple}{0.2$\uparrow$}/\textcolor{purple}{0.5$\uparrow$} in image-level scores and \textcolor{purple}{0.4$\uparrow$}/\textcolor{purple}{4.1$\uparrow$}/\textcolor{purple}{2.9$\uparrow$}/\textcolor{purple}{0.9$\uparrow$} in pixel-level scores. Furthermore, on the more challenging Real-IAD dataset, INP-Former++ achieves image level scores of \textbf{90.7}/\textbf{88.3}/\textbf{81.6} and pixel level scores of \textbf{99.2}/\textbf{50.2}/\textbf{50.9}/\textbf{95.1}, exceeding the second-best method, Dinomaly, by \textcolor{purple}{1.3$\uparrow$}/\textcolor{purple}{1.5$\uparrow$}/\textcolor{purple}{1.4$\uparrow$} in image-level scores and \textcolor{purple}{0.2$\uparrow$}/\textcolor{purple}{10.4$\uparrow$}/\textcolor{purple}{6.0$\uparrow$}/\textcolor{purple}{1.2$\uparrow$} in pixel-level scores. These results further highlight the robustness and effectiveness of INP-Former++ in handling complex real-world scenarios. Compared with our conference version, INP-Former, the enhanced INP-Former++ achieves a slight improvement in image-level scores across all three datasets while exhibiting a substantial boost in pixel-level performance.  Specifically, it improves pixel-level scores by \textcolor{purple}{0.2$\uparrow$}/\textcolor{purple}{7.6$\uparrow$}/\textcolor{purple}{3.9$\uparrow$}/\textcolor{purple}{1.3$\uparrow$} on MVTec-AD, \textcolor{purple}{0.1$\uparrow$}/\textcolor{purple}{5.9$\uparrow$}/\textcolor{purple}{4.1$\uparrow$}/\textcolor{purple}{1.0$\uparrow$} on VisA, and \textcolor{purple}{0.1$\uparrow$}/\textcolor{purple}{6.0$\uparrow$}/\textcolor{purple}{3.0$\uparrow$}/\textcolor{purple}{0.2$\uparrow$} on Real-IAD.\\
\indent Fig.~\ref{fig:anomalymap} illustrates the qualitative results of anomaly localization obtained by INP-Former and INP-Former++. While INP-Former can identify anomalous regions to some extent, its outputs often suffer from considerable background noise. In contrast, the improved INP-Former++ not only achieves more accurate anomaly localization but also effectively suppresses background interference. This improvement can be attributed to the proposed residual learning strategy, which effectively amplifies the residual differences between normal and anomalous regions.

\begin{table*}[]
\centering
\caption{\textbf{Super-multi-class} anomaly detection performance on different AD datasets. $\Delta$ represents the performance change of \textbf{INP-Former++} in the super-multi-class setting compared with the multi-class setting.}
\label{table:super-multi-class}
\vspace{-10pt}
\fontsize{10}{14}\selectfont{
\resizebox{\linewidth}{!}{
\begin{tabular}{c|cccccc}
\toprule[1.5pt]
Dataset~$\rightarrow$           & \multicolumn{2}{c|}{MVTec-AD~\cite{MVTec-AD}}                             & \multicolumn{2}{c|}{VisA~\cite{VisA}}                                 & \multicolumn{2}{c}{Real-IAD~\cite{real-iad}}         \\ \midrule
Setting~$\downarrow$           & Image-level    & \multicolumn{1}{c|}{Pixel-level}         & Image-level    & \multicolumn{1}{c|}{Pixel-level}         & Image-level    & Pixel-level         \\ \midrule
Multi-Class       & 99.8/99.9/99.3 & \multicolumn{1}{c|}{98.7/75.6/71.6/96.0} & 98.9/99.0/96.4 & \multicolumn{1}{c|}{99.1/52.7/55.2/96.0} & 90.7/88.3/81.6 & 99.2/50.2/50.9/95.1 \\
Super-Multi-Class & 99.6/99.8/99.0 & \multicolumn{1}{c|}{98.5/73.9/70.8/95.5} & 97.8/98.1/94.5 & \multicolumn{1}{c|}{98.9/53.4/54.8/94.6} & 90.6/88.3/81.4 & 99.1/49.6/50.5/95.0 \\ \midrule
$\Delta$                 & \textcolor{purple}{0.2$\downarrow$/0.1$\downarrow$/0.3$\downarrow$}    & \multicolumn{1}{c|}{\textcolor{purple}{0.2$\downarrow$/1.7$\downarrow$/0.8$\downarrow$/0.5$\downarrow$}}     & \textcolor{purple}{1.1$\downarrow$/0.9$\downarrow$/1.9$\downarrow$}    & \multicolumn{1}{c|}{\textcolor{purple}{0.2$\downarrow$/0.7$\uparrow$/0.4$\downarrow$/1.4$\downarrow$}}   & \textcolor{purple}{0.1$\downarrow$/0.0/0.2$\downarrow$}    & \textcolor{purple}{0.1$\downarrow$/0.6$\downarrow$/0.4$\downarrow$/0.1$\downarrow$}     \\ \bottomrule[1.5pt]
\end{tabular}}}
\end{table*}
\subsubsection{Super-Multi-Class Anomaly Detection}
To further evaluate the performance of INP-Former++, we jointly train the model on three datasets—MVTec-AD, VisA, and Real-IAD—encompassing a total of 57 categories, thereby establishing a super-multi-class anomaly detection setting.\\
\indent As reported in Tab.~\ref{table:super-multi-class}, compared to the multi-class setting, our method exhibits only a marginal performance drop under the super-multi-class setting. These results demonstrate that INP-Former++ exhibits superior scalability, allowing a single unified model to detect a wider range of product categories, thereby offering significant potential for reducing memory overhead in industrial applications.

\begin{table*}[!ht]
\centering
\caption{\textbf{Few-shot (4-shot)} anomaly detection performance on different AD datasets. The best in \textbf{bold}, the
second-highest is \underline{underlined}. $\dagger$ indicates the results we reproduced using publicly available code. ``-" denotes not available.}
\label{table:few-shot-main-performance}
\vspace{-10pt}
\fontsize{10}{14}\selectfont{
\resizebox{1.0\linewidth}{!}{
\begin{tabular}{c|cc|cc|cc}
\toprule[1.5pt]
Dataset~$\rightarrow$    & \multicolumn{2}{c|}{MVTec-AD~\cite{MVTec-AD}}        & \multicolumn{2}{c|}{VisA~\cite{VisA}}              & \multicolumn{2}{c}{Real-IAD~\cite{real-iad}}                                                      \\ \midrule
Method~$\downarrow$     & Image-level    & Pixel-level         & Image-level    & Pixel-level           & \multicolumn{1}{c}{Image-level}                            & Pixel-level         \\ \midrule
SPADE~\cite{SPADE}      & 84.8/92.5/91.5 & 92.7/-/46.2/87.0    & 81.7/83.4/82.1 & 96.6/-/43.6/87.3      & \multicolumn{1}{c}{50.8$^{\dagger}$/45.8$^{\dagger}$/61.2$^{\dagger}$}                            & 59.5$^{\dagger}$/0.2$^{\dagger}$/0.5$^{\dagger}$/19.2$^{\dagger}$         \\
PaDiM~\cite{defard2021padim}      & 80.4/90.5/90.2 & 92.6/-/46.1/81.3    & 72.8/75.6/78.0 & 93.2/-/24.6/72.6      & \multicolumn{1}{c}{60.3$^{\dagger}$/53.5$^{\dagger}$/64.0$^{\dagger}$}                            & 90.9$^{\dagger}$/2.1$^{\dagger}$/5.1$^{\dagger}$/67.6$^{\dagger}$         \\
PatchCore~\cite{Patchcore}  & 88.8/94.5/92.6 & 94.3/-/55.0/84.3    & 85.3/87.5/84.3 & 96.8/-/43.9/84.9      & \multicolumn{1}{c}{66.0$^{\dagger}$/62.2$^{\dagger}$/65.2$^{\dagger}$}                            & 92.9$^{\dagger}$/9.8$^{\dagger}$/16.1$^{\dagger}$/68.6$^{\dagger}$         \\
RegAD~\cite{regad}  & 88.2/94.8/- & 95.8/-/-/88.1    & 73.8/75.8/- & 95.9/-/-/76.5      & \multicolumn{1}{c}{-/-/-} & -/-/-/- \\
WinCLIP~\cite{WinClip}    & 95.2/97.3/94.7 & 96.2/-/59.5/89.0    & 87.3/88.8/84.2 & 97.2/-/47.0/87.6      & \multicolumn{1}{c}{73.0$^{\dagger}$/61.8$^{\dagger}$/61.0$^{\dagger}$}                         & 93.8$^{\dagger}$/13.3$^{\dagger}$/21.0$^{\dagger}$/76.4$^{\dagger}$ \\
PromptAD~\cite{PromptAD}   & 96.6/-/-       & 96.5/-/-/90.5       & 89.1/-/-       & 97.4/-/-/86.2         & \multicolumn{1}{c}{59.7$^{\dagger}$/43.5$^{\dagger}$/52.9$^{\dagger}$}                         & 86.9$^{\dagger}$/8.7$^{\dagger}$/16.1$^{\dagger}$/61.9$^{\dagger}$  \\ 
SimCLIP~\cite{deng2024simclip}   & 96.4/98.0/-       & 96.2/-/-/\underline{93.1}       & 94.4/95.6/-       & \underline{98.0}/-/-/94.1         & \multicolumn{1}{c}{-/-/-}  &  -/-/-/- \\\midrule
\textbf{INP-Former}\footref{reproduce}~\cite{luo2025exploring} & \underline{97.6}/\underline{98.6}/\textbf{97.0} & \underline{97.0}/\underline{62.9}/\underline{63.8}/92.7& \underline{96.4}/\underline{96.0}/\underline{93.0} & 97.8/\underline{44.2}/\underline{50.7}/\underline{94.3} & \multicolumn{1}{c}{\underline{76.7}/\underline{72.3}/\underline{71.7}} & \underline{97.5}/\underline{30.1}/\underline{35.1}/\underline{89.6} \\
\rowcolor{Light} 
\textbf{INP-Former++} & \textbf{97.9}/\textbf{98.9}/\underline{96.9} & \textbf{97.3}/\textbf{69.8}/\textbf{67.6}/\textbf{94.0}& \textbf{96.7}/\textbf{96.1}/\textbf{93.6} & \textbf{98.3}/\textbf{48.9}/\textbf{53.6}/\textbf{95.2} & \multicolumn{1}{c}{\cellcolor{Light}\textbf{77.2}/\textbf{72.5}/\textbf{72.3}} & \textbf{97.7}/\textbf{35.5}/\textbf{38.0}/\textbf{90.2} \\\bottomrule[1.5pt]
\end{tabular}}}
\vspace{-3mm}
\end{table*}
\subsubsection{Few-Shot Anomaly Detection}
Under the few-shot setting,  we compare INP-Former++ with a range of SOTA methods, including prototype-based methods such as SPADE~\cite{SPADE}, PaDiM~\cite{defard2021padim}, and PatchCore~\cite{Patchcore}; spatial alignment-based methods like RegAD~\cite{regad} , and recent VLM-based methods, such as WinCLIP~\cite{WinClip}, PromptAD~\cite{PromptAD}, and SimCLIP~\cite{deng2024simclip}.\\
\indent As shown in Tab.~\ref{table:few-shot-main-performance}, INP-Former++ substantially outperforms existing methods specifically designed for few-shot anomaly detection, achieving SOTA performance on three industrial AD datasets. On MVTec-AD, it achieves image-level scores of \textbf{97.9}/\textbf{98.9}/\underline{96.9} and pixel-level scores of \textbf{97.3}/\textbf{69.8}/\textbf{67.6}/\textbf{94.0}. On VisA, it obtains image-level scores of \textbf{96.7}/\textbf{96.1}/\textbf{93.6} and pixel-level scores of \textbf{98.3}/\textbf{48.9}/\textbf{53.6}/\textbf{95.2}. On the more challenging Real-IAD, it reaches image-level scores of \textbf{77.2}/\textbf{72.5}/\textbf{72.3} and pixel-level scores of \textbf{97.7}/\textbf{35.5}/\textbf{38.0}/\textbf{90.2}. These strong results can be attributed to the ability of our method to extract INPs from a single image, thereby mitigating the need for large-scale training data. Compared with our earlier conference version, INP-Former, the enhanced INP-Former++ achieves consistent improvements in both image-level and pixel-level performance. For example, on Real-IAD, it improves the image-level scores by \textcolor{purple}{0.5$\uparrow$}/\textcolor{purple}{0.2$\uparrow$}/\textcolor{purple}{0.6$\uparrow$} and the pixel-level scores by \textcolor{purple}{0.2$\uparrow$}/\textcolor{purple}{5.4$\uparrow$}/\textcolor{purple}{2.9$\uparrow$}/\textcolor{purple}{0.6$\uparrow$}. These gains are primarily due to the introduction of the Soft INP Coherence Loss, which encourages the model to learn more representative INPs.

\begin{table*}[!ht]
\centering
\caption{\textbf{Single class} anomaly detection performance on different AD datasets. The best in \textbf{bold}, the
second-highest is \underline{underlined}. ``-" denotes not available.}
\label{table:single-class-main-performance}
\vspace{-10pt}
\fontsize{11.5}{14}\selectfont{
\resizebox{.85\textwidth}{!}{
\begin{tabular}{c|ccc|ccc|ccc}
\toprule[1.5pt]
Dataset~$\rightarrow$     & \multicolumn{3}{c|}{MVTec-AD~\cite{MVTec-AD}} & \multicolumn{3}{c|}{VisA~\cite{VisA}} & \multicolumn{3}{c}{Real-IAD~\cite{real-iad}} \\ \midrule
Method~$\downarrow$      & I-AUROC    & P-AP   & AUPRO   & I-AUROC  & P-AP  & AUPRO  & I-AUROC   & P-AP   & AUPRO   \\ \midrule
PatchCore~\cite{Patchcore}   & 99.1       & 56.1   & 93.5    & 95.1     & 40.1  & 91.2   & 89.4      & -      & 91.5    \\
RD4AD~\cite{RD4AD}       & 98.5       & 58.0     & 93.9    & 96.0       & 27.7  & 70.9   & 87.1      & -      & 93.8    \\
SimpleNet~\cite{liu2023simplenet}   & 99.6       & 54.8   & 90.0      & 96.8     & 36.3  & 88.7   & 88.5      & -      & 84.6    \\
Dinomaly\footref{reproduce}~\cite{guo2024dinomaly}    & \underline{99.7}       & {66.1}   & {94.6}      & \underline{98.9}     & \underline{45.6}  & \underline{95.3}   & 92.0        & 42.0   & 95.1    \\ \midrule
\textbf{INP-Former}\footref{reproduce}~\cite{luo2025exploring}  & \underline{99.7}       & \underline{67.0}   & \underline{95.4}    & {98.5}        & {44.3}     & {94.7}      & \underline{92.1}      & \underline{44.1}   & \underline{95.8}    \\ 
\rowcolor{Light} 
\textbf{INP-Former++}  & \textbf{99.8}       & \textbf{76.6}   & \textbf{96.4}    & \textbf{99.0}        & \textbf{51.1}     & \textbf{95.7}      & \textbf{92.5}      & \textbf{54.9}   & \textbf{95.9}    \\\bottomrule[1.5pt]
\end{tabular}}
}
\vspace{-3mm}
\end{table*}
\subsubsection{Single-Class Anomaly Detection}
We also compare INP-Former++ with existing SOTA methods in the single-class setting, as shown in Tab.~\ref{table:single-class-main-performance}. The results indicate that our model achieves SOTA performance across all three datasets. For instance, on the widely used MVTec-AD dataset, INP-Former++ achieves an I-AUROC of \textbf{99.8}, P-AP of \textbf{76.6}, and AUPRO of \textbf{96.4}, surpassing the second-best method, Dinomaly, by \textcolor{purple}{0.1$\uparrow$}, \textcolor{purple}{10.5$\uparrow$}, and \textcolor{purple}{1.8$\uparrow$}, respectively. Compared with our conference version, INP-Former++ shows improvements of \textcolor{purple}{0.1$\uparrow$} in I-AUROC, \textcolor{purple}{9.6$\uparrow$} in P-AP, and \textcolor{purple}{1.0$\uparrow$} in AUPRO.


\begin{table*}[]
\centering
\caption{\textbf{Semi-supervised (10 anomalous samples per category)} anomaly detection performance on different AD datasets. The best in \textbf{bold}, the
second-highest is \underline{underlined}. To ensure fairness, \textbf{we remove the added abnormal samples from the test set} and reproduce the performance of the compared methods in the \textbf{multi-class} setting based on their publicly available implementations. $*$ denotes that we observe a significant performance degradation of BGAD under the multi-class setting.}
\label{table:semi-supervised-main-performance}
\vspace{-10pt}
\fontsize{10}{14}\selectfont{
\resizebox{\textwidth}{!}{
\begin{tabular}{c|cc|cc|cc}
\toprule[1.5pt]
Dataset~$\rightarrow$        & \multicolumn{2}{c|}{MVTec-AD~\cite{MVTec-AD}}        & \multicolumn{2}{c|}{VisA~\cite{VisA}}             & \multicolumn{2}{c}{Real-IAD~\cite{real-iad}}         \\ \midrule
Method~$\downarrow$         & Image-level    & Pixel-level         & Image-level    & Pixel-level          & Image-level    & Pixel-level         \\ \midrule
CDO~\cite{CDO}            & 97.5/98.5/96.4 & \underline{98.7}/\underline{74.3}/\underline{69.9}/\underline{95.0} & 86.4/88.0/83.5 & \underline{98.9}/\underline{48.4}/\underline{50.3}/\underline{91.3}  & 77.7/75.1/70.4 & 97.0/28.6/33.6/84.9 \\
DeSTSeg~\cite{zhang2023destseg}        & 96.6/98.8/96.4 & 90.9/63.2/63.5/80.0 & \underline{92.7}/\underline{94.4}/\underline{88.8} & 90.9/43.0/48.7/68.7  & 81.9/\underline{81.5}/73.0 & 83.7/27.8/34.7/64.1 \\
RD++~\cite{RDplus}           & \underline{98.6}/\underline{99.4}/\underline{97.5} & 96.7/53.2/57.3/92.6 & 91.8/90.9/88.7 & 98.6/42.1/46.4/91.2  & \underline{83.0}/{78.3}/\underline{74.7} & \underline{97.9}/23.9/31.6/\underline{90.2} \\
BGAD$^*$~\cite{BGAD}           & 76.4/86.9/86.6 & 95.1/41.8/44.5/87.8 &58.8/62.4/71.3          & 91.2/6.5/10.2/68.0              & 52.7/48.7/61.3          & 90.5/3.3/6.4/72.1             \\
SuperSimpleNet~\cite{rolih2025supersimplenet} & 94.0/97.4/94.4 & 97.6/71.8/68.3/93.1 & 82.6/85.8/80.3 & 97.3/44.7/47.0/87.4 & 71.9/69.5/66.4         & 94.9/\underline{29.7}/\underline{35.2}/81.6             \\ \midrule
\rowcolor{Light} 
\textbf{INP-Former++}   & \textbf{99.7}/\textbf{99.8}/\textbf{99.0} & \textbf{99.1}/\textbf{79.6}/\textbf{75.2}/\textbf{96.8} & \textbf{99.0}/\textbf{99.0}/\textbf{96.5} & \textbf{99.2}/\textbf{57.2}/\textbf{58.6}/\textbf{96.4}  & \textbf{91.6}/\textbf{89.4}/\textbf{82.9} & \textbf{99.2}/\textbf{55.2}/\textbf{55.9}/\textbf{95.4} \\ \bottomrule[1.5pt]
\end{tabular}}}
\end{table*}
\subsubsection{Semi-Supervised Anomaly Detection}
In industrial scenarios, only a limited number of anomalous samples are typically available, making it crucial to effectively incorporate the knowledge from these anomalies. Therefore, we conduct performance comparison experiments under a semi-supervised setting, where each category includes 10 anomalous samples. Considering the memory constraints commonly encountered in industrial applications, our semi-supervised experiments are performed under a multi-class setting. We reproduce the semi-supervised performance of CDO\footnote{\href{https://github.com/caoyunkang/CDO}{https://github.com/caoyunkang/CDO}}~\cite{CDO}, DeSTSeg\footnote{\href{https://github.com/apple/ml-destseg}{https://github.com/apple/ml-destseg}}~\cite{zhang2023destseg}, RD++\footnote{\href{https://github.com/tientrandinh/Revisiting-Reverse-Distillation}{https://github.com/tientrandinh/Revisiting-Reverse-Distillation}}~\cite{RDplus}, BGAD\footnote{\href{https://github.com/xcyao00/BGAD}{https://github.com/xcyao00/BGAD}}~\cite{BGAD}, and SuperSimpleNet\footnote{\href{https://github.com/blaz-r/SuperSimpleNet}{https://github.com/blaz-r/SuperSimpleNet}}~\cite{rolih2025supersimplenet} using their publicly available official implementations.\\
\indent As shown in Tab.~\ref{table:semi-supervised-main-performance}, the proposed INP-Former++ achieves SOTA performance under the semi-supervised setting, substantially outperforming existing methods. For instance, on the MVTec-AD dataset, INP-Former++ surpasses the second-best results by \textcolor{purple}{1.1$\uparrow$}/\textcolor{purple}{0.4$\uparrow$}/\textcolor{purple}{1.5$\uparrow$} in image-level scores and by \textcolor{purple}{0.4$\uparrow$}/\textcolor{purple}{5.3$\uparrow$}/\textcolor{purple}{5.3$\uparrow$}/\textcolor{purple}{1.8$\uparrow$} in pixel-level scores. Compared to its performance under the unsupervised multi-class setting (Tab.~\ref{table:multi-class-main-performance}), the anomaly localization accuracy is significantly enhanced, For example, the pixel-level scores on the MVTec-AD dataset improve by \textcolor{purple}{0.4$\uparrow$}/\textcolor{purple}{4.0$\uparrow$}/\textcolor{purple}{3.6$\uparrow$}/\textcolor{purple}{0.8$\uparrow$}. This improvement is primarily attributed to the designed residual learning strategy, which effectively integrates knowledge from a limited number of real anomalies into the model.


\begin{table}[]
\centering
\caption{\textbf{Multi-class} anomaly detection performance on \textbf{Uni-Medical}~\cite{zhang2023exploring} dataset with image-level (I-AUROC/I-AP/I-F1\_max) and pixel-level (P-AUROC/P-AP/P-F1\_max/AUPRO) metrics. The best in \textbf{bold}, the
second-highest is \underline{underlined}.}
\label{table:medical}
\vspace{-10pt}
\fontsize{10.5}{14}\selectfont{
\resizebox{0.925\linewidth}{!}{
\begin{tabular}{c|cc}
\toprule[1.5pt]
Method~$\downarrow$       & Image-level    & Pixel-level         \\ \midrule
RD4AD~\cite{RD4AD}        & 76.1/75.3/78.2 & 96.5/38.3/39.8/\underline{86.8} \\
UniAD~\cite{uniad}        & 79.0/76.1/77.1 & 96.6/39.3/41.1/86.0 \\
SimpleNet~\cite{liu2023simplenet}    & 77.5/77.7/76.7 & 94.3/34.4/36.0/77.0 \\
DeSTSeg~\cite{zhang2023destseg}      & 78.5/77.0/78.2 & 65.7/41.7/34.0/35.3 \\
DiAD~\cite{diad}         & 78.8/77.2/77.7 & 95.8/34.2/35.5/84.3 \\
MambaAD~\cite{he2024mambaad}      & \underline{83.9}/80.8/\textbf{81.9} & \underline{96.8}/45.8/47.5/\textbf{88.2} \\
Dinomaly\footref{reproduce}~\cite{guo2024dinomaly}     & 83.4/\underline{83.1}/80.6 & 96.7/\underline{50.4}/\underline{50.6}/84.8 \\ \midrule
\rowcolor{Light}
\textbf{INP-Former++} & \textbf{86.7}/\textbf{85.6}/\underline{81.8} & \textbf{97.0}/\textbf{52.5}/\textbf{54.2}/86.2 \\ \bottomrule[1.5pt]
\end{tabular}}}
\end{table}
\subsubsection{Medical Anomaly Detection}
To further validate the generalizability of the proposed INP-Former++, we not only conduct experiments on industrial AD datasets but also extend its application to the medical AD domain. Tab.~\ref{table:medical} presents the quantitative comparison results in a multi-class setting within the medical domain. INP-Former++ demonstrates outstanding performance, achieving image-level scores of \textbf{86.7}/\textbf{85.6}/\textbf{81.8} and pixel-level scores of \textbf{97.0}/\textbf{52.5}/\textbf{54.2}/86.2. Compared to the second-best results, the I-AUROC/I-AP improve by \textcolor{purple}{2.8$\uparrow$}/\textcolor{purple}{2.5$\uparrow$}, and P-AUROC/P-AP/P-F1\_max increase by \textcolor{purple}{0.2$\uparrow$}/\textcolor{purple}{2.1$\uparrow$}/\textcolor{purple}{3.6$\uparrow$}.  These results highlight the excellent performance of INP-Former++ in medical domain, further confirming the domain generalizability of our method.

\begin{table*}[]
\centering
\caption{\textbf{Overall ablation} on MVTec-AD~\cite{MVTec-AD} and VisA~\cite{VisA} datasets. \textbf{\textit{``INP''}} refers to the use of INP Extractor and INP-guided Decoder. \textbf{\textit{``RL''}} denotes the Residual Learning strategy. {\textbf{\textit{``RL''}}$^{\ddagger}$ denotes the Residual Learning strategy \textit{\textbf{without stop-gradient}} operation.}}
\label{table:ablation}
\vspace{-10pt}
\fontsize{10}{14}\selectfont{
\resizebox{\linewidth}{!}{
\begin{tabular}{c|cccccc|cc|cc}
\toprule[1.5pt]
                        & \multicolumn{6}{c|}{Dataset~$\rightarrow$} & \multicolumn{2}{c|}{MVTec-AD~\cite{MVTec-AD} }              & \multicolumn{2}{c}{VisA~\cite{VisA} }                         \\ \midrule
                        & \textbf{\textit{``INP''}}    & $\mathcal{L}_c$    & $\mathcal{L}_{sc}$    & $\mathcal{L}_{sm}$   & \textbf{\textit{``RL''}}   & \textbf{\textit{``RL''}}$^{\ddagger}$ & Image-level       & Pixel-level             & Image-level       & Pixel-level               \\ \midrule
\multirow{7}{*}{\rotatebox{90}{\textbf{Module}}} & \xmark     & \xmark   & \xmark   & \xmark & \xmark  & \xmark   & 98.53/99.13/97.45 & 97.05/57.28/60.69/92.87 & 96.86/97.41/93.25 & 97.58/42.18/48.18/94.04          \\
                        & \cmark     & \xmark   & \xmark   & \xmark  & \xmark  & \xmark & 99.54/99.80/98.93                   & 98.25/65.32/65.86/94.52                        & 98.15/98.21/95.22                  & 98.81/46.94/50.90/95.38                                             \\
                        & \cmark     & \cmark   & \xmark   & \xmark  & \xmark & \xmark  & 99.61/99.81/99.01                   &  98.28/65.54/66.21/94.63                       & 98.21/98.28/95.32                  & 98.85/47.35/51.42/95.43 \\
                        & \cmark     & \xmark   & \cmark   & \xmark  & \xmark & \xmark  & 99.69/99.85/99.21                   & 98.30/65.88/66.60/94.87                         & 98.23/98.30/95.35                  & 98.90/47.83/51.49/95.47                                            \\
                        & \cmark     & \xmark   & \cmark   & \cmark  & \xmark  & \xmark & 99.76/99.90/\textbf{99.31}                   &98.46/67.01/67.26/94.95                         & 98.86/98.99/96.30                  &  99.08/47.49/51.65/95.68                                          \\
                        & \cellcolor{Light}\cmark     & \cellcolor{Light}\xmark   & \cellcolor{Light}\cmark   & \cellcolor{Light}\cmark  & \cellcolor{Light}\cmark   & \cellcolor{Light}\xmark & \cellcolor{Light}\textbf{99.78}/\textbf{99.91}/99.26                 & \cellcolor{Light}\textbf{98.71}/\textbf{75.56}/\textbf{71.62}/\textbf{95.98}                       & \cellcolor{Light} \textbf{98.91}/\textbf{99.00}/\textbf{96.43}                 &   \cellcolor{Light}  \textbf{99.10}/\textbf{52.67}/\textbf{55.17}/\textbf{95.99} \\ 
                         & \cmark     & \xmark   & \cmark   & \cmark  & \xmark  & \cmark & 99.70/99.84/99.25                   &96.57/63.37/63.67/90.68                         & 98.84/98.97/96.28                  &  99.02/48.24/51.65/95.46                                          \\ \bottomrule[1.5pt]
\end{tabular}}}
\end{table*}
\begin{figure}
    \centering
    \includegraphics[width=\linewidth]
    {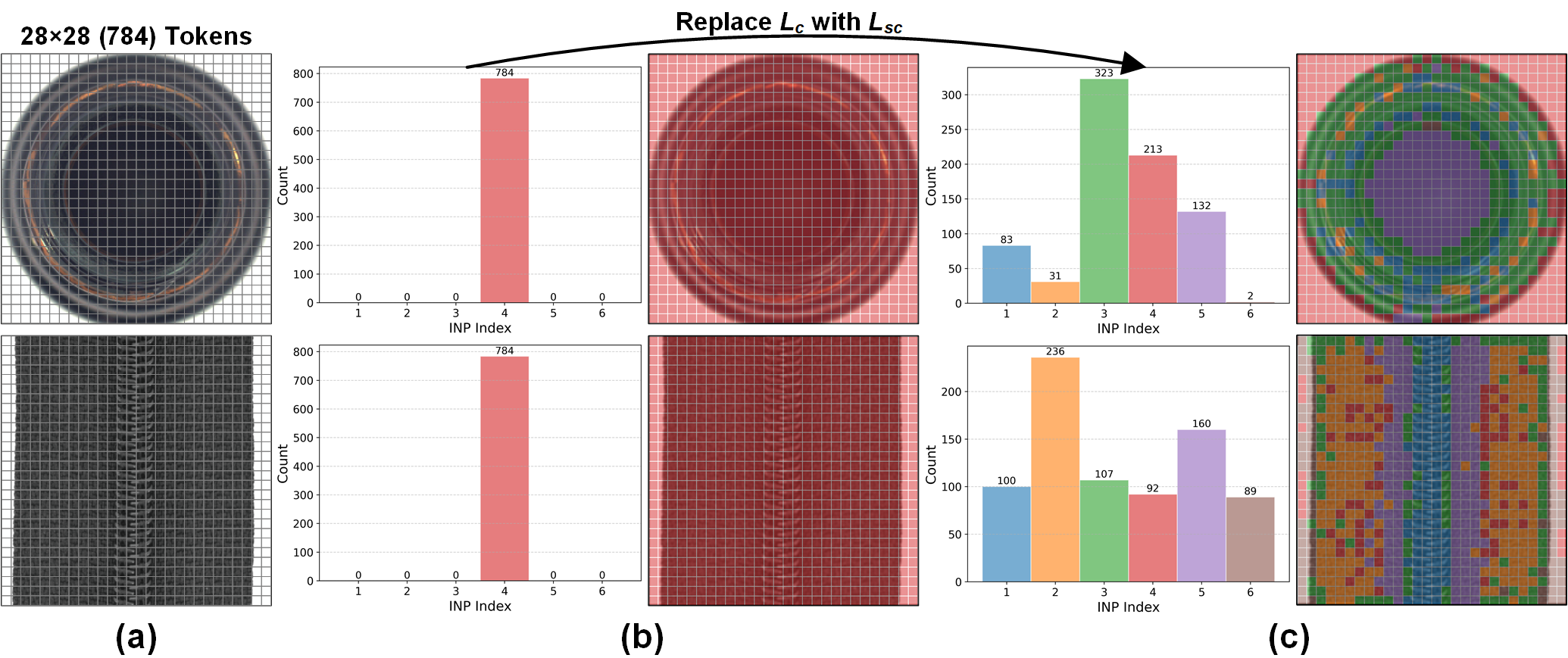}
    \vspace{-20pt}
    \caption{\textbf{Shortcut solution induced by $\mathcal{L}_{c}$.} (a) The input image is divided into 784 tokens. (b) \textbf{With $\mathcal{L}_{c}$}, the INP extractor collapses to a shortcut, assigning all tokens to the same INP. (c) \textbf{With $\mathcal{L}_{sc}$}, tokens with different semantics are assigned to different INPs.}
    \label{fig:lc_shortcut}
\end{figure}

\begin{figure}[!t]
    \centering
    \includegraphics[width=\linewidth]{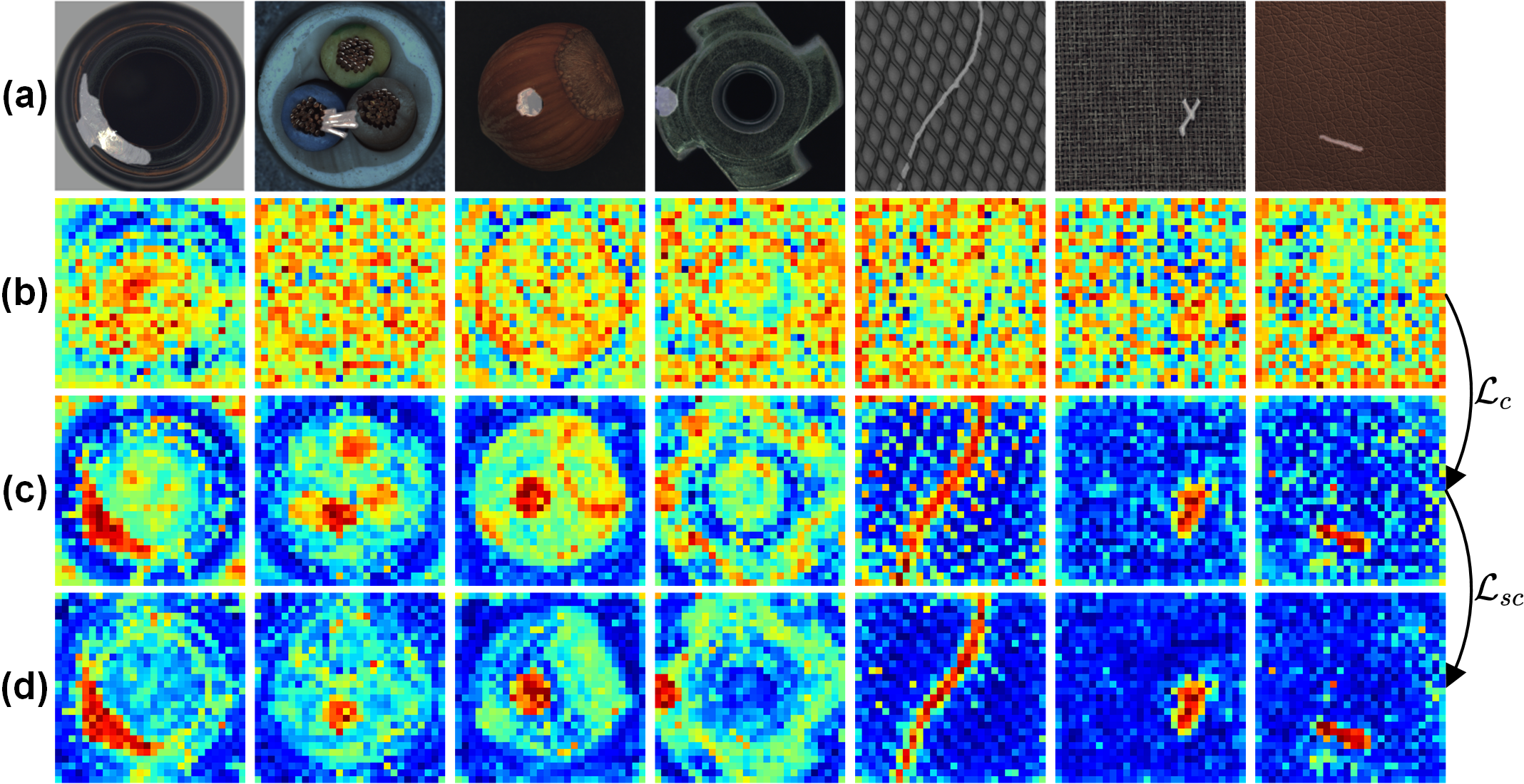}
    \vspace{-20pt}
    \caption{\textbf{Visualization of the impact of soft INP coherence loss $\mathcal{L}_{sc}$}. (a) Input anomalous image and ground truth. (b) Distance map \textbf{without $\mathcal{L}_{c}$}. (c) Distance map \textbf{with $\mathcal{L}_{c}$}. (d) Distance map \textbf{with $\mathcal{L}_{sc}$}.}
    \label{fig:ablation_lsc}
\end{figure}

\begin{figure}[!t]
    \centering
    \includegraphics[width=\linewidth]{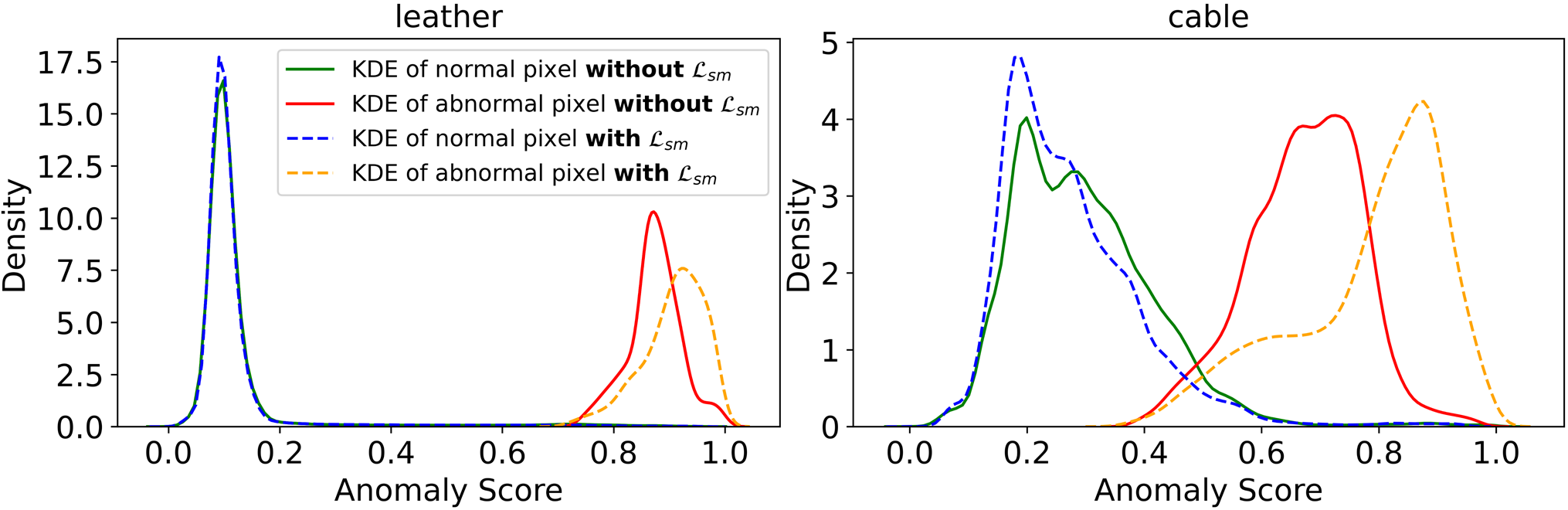}
    \vspace{-20pt}
    \caption{\textbf{Visualization of the impact of soft mining loss $\mathcal{L}_{sm}$}. We plot the Kernel Density Estimation (KDE) for the leather and cable categories in the MVTec-AD~\cite{MVTec-AD} dataset to estimate the probability density of the anomaly scores.}
    \label{fig:ablation_lsm}
\end{figure}

\begin{figure}[!t]
    \centering
    \includegraphics[width=\linewidth]{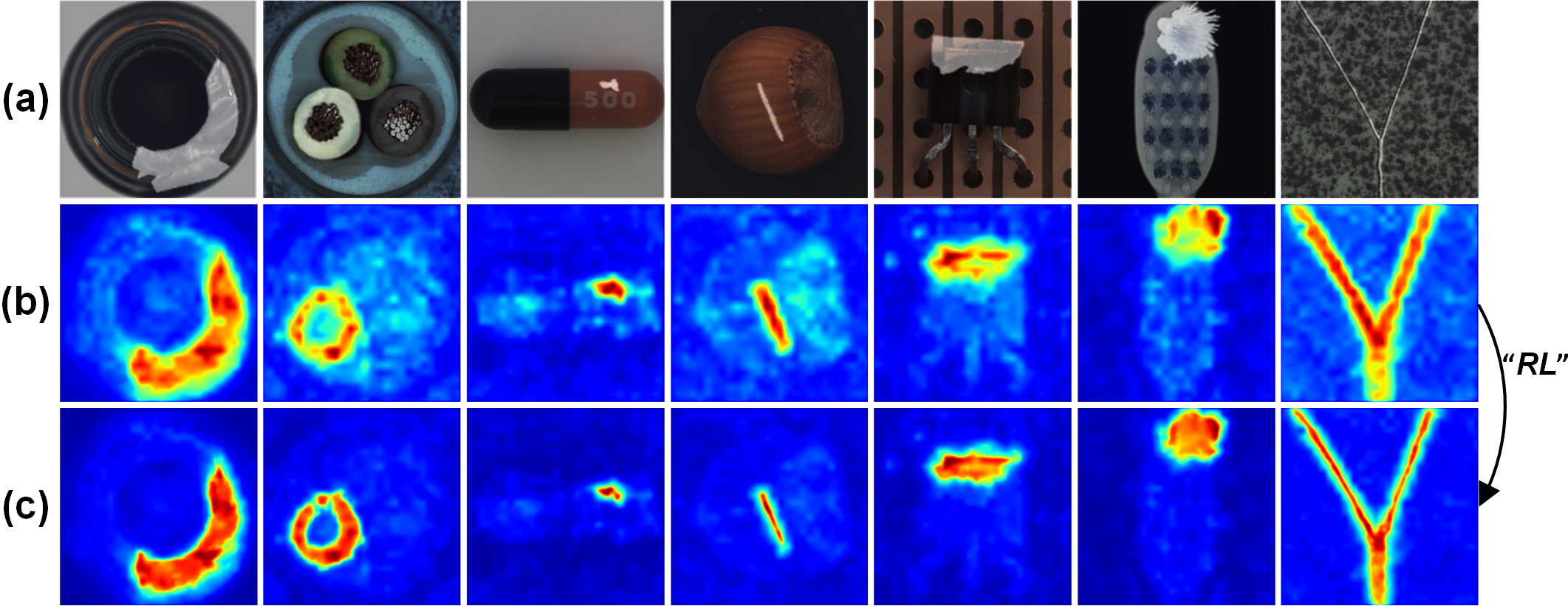}
    \vspace{-20pt}
    \caption{\textbf{Visualization of the impact of residual learning (\textbf{\textit{``RL''}})}. (a) Input anomalous image and ground truth. (b) Predicted anomaly map \textbf{without \textbf{\textit{``RL''}}}. (c) Predicted anomaly map \textbf{with \textbf{\textit{``RL''}}}.}
    \label{fig:ablation_rl}
\end{figure}

\begin{figure*}[!t]
    \centering
    \includegraphics[width=0.98\linewidth]{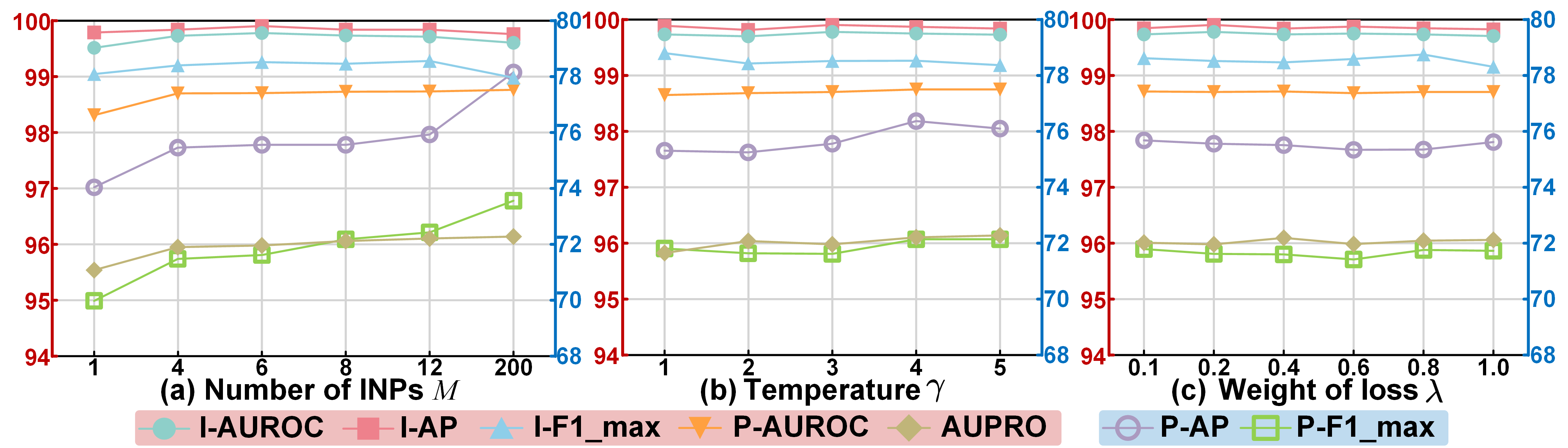}
    \vspace{-12pt}
    \caption{\textbf{Ablation analysis of model hyperparameters on the MVTec-AD~\cite{MVTec-AD} dataset.} (a) The number of INPs $M$. (b) The temperature parameter $\gamma$ in $\mathcal{L}_{sm}$, (c) The loss weight $\lambda$.}
    \label{fig:ablation-hyber}
\end{figure*}
\subsection{Ablation Study}
\subsubsection{Overall Ablation}
\label{sec-overall-ablation}
As presented in Tab.~\ref{table:ablation}, we conduct comprehensive ablation studies on the MVTec-AD~\cite{MVTec-AD} and VisA~\cite{VisA} datasets to evaluate the effectiveness of each proposed component, \ie, INP Extractor and INP-guided Decoder (\textbf{\textit{``INP''}}), INP Coherence Loss ($\mathcal{L}_{c}$), Soft INP coherence Loss ($\mathcal{L}_{sc}$), Soft Mining Loss ($\mathcal{L}_{sm}$), and Residual Learning Strategy (\textbf{\textit{``RL''}}).\\
\indent In the first row, we train a baseline model without incorporating any of the proposed components, similar to RD4AD~\cite{RD4AD} framework. The results in the second row clearly validate the effectiveness of \textbf{\textit{``INP''}}, which substantially enhances the model performance. Specifically, on the MVTec-AD dataset, the image-level and pixel-level scores improve by \textcolor{purple}{1.01$\uparrow$}/\textcolor{purple}{0.67$\uparrow$}/\textcolor{purple}{1.48$\uparrow$} and \textcolor{purple}{1.20$\uparrow$}/\textcolor{purple}{8.04$\uparrow$}/\textcolor{purple}{5.17$\uparrow$}/\textcolor{purple}{1.65$\uparrow$}, respectively. On the VisA dataset, the corresponding improvements are \textcolor{purple}{1.29$\uparrow$}/\textcolor{purple}{0.80$\uparrow$}/\textcolor{purple}{1.97$\uparrow$} and \textcolor{purple}{1.23$\uparrow$}/\textcolor{purple}{4.76$\uparrow$}/\textcolor{purple}{2.72$\uparrow$}/\textcolor{purple}{1.34$\uparrow$}. These improvements are attributed to the information bottleneck introduced by \textbf{\textit{``INP''}}: the INP Extractor compresses semantic information from the input features into a few representative INPs, which are then used by the INP-guided Decoder to reconstruct the features. This process helps the model preserve normal features while filtering out anomalous ones.\\
\indent The results in the third row indicate that $\mathcal{L}_{c}$ further improves the model performance. This improvement stems from $\mathcal{L}_{c}$ ensuring that the extracted INP
coherently represents normal patterns, thereby avoiding the capture of anomalous ones and establishing a solid foundation for the subsequent suppression of anomalous feature reconstruction. The comparison between (b) and (c) in Fig.~\ref{fig:ablation_lsc} provides a more intuitive demonstration of the effectiveness of $\mathcal{L}_{c}$.\\ 
\indent However, as shown in Fig.~\ref{fig:lc_shortcut}(b), we observe that $\mathcal{L}_{c}$ tends to induce a shortcut solution in the INP Extractor, where all patch tokens are incorrectly assigned to the same INP. To overcome this limitation, we propose $\mathcal{L}_{sc}$ as a replacement for $\mathcal{L}_{c}$. As illustrated in Fig.~\ref{fig:lc_shortcut}(c), $\mathcal{L}_{sc}$ encourages semantically distinct patch tokens to be assigned to different INPs, thereby more effectively preserving the underlying semantic structure. Additionally, the quantitative results in the fourth row of Tab.~\ref{table:ablation} provide additional evidence of the superiority of $\mathcal{L}_{sc}$ over $\mathcal{L}_{c}$. The comparison between Figs.~\ref{fig:ablation_lsc}(c) and \ref{fig:ablation_lsc}(d) offers a more intuitive demonstration of this advantage. Furthermore, Fig.~\ref{fig:anomalymapZS} and Tab.~\ref{table:zero-shot} qualitatively and quantitatively validate the superiority of $\mathcal{L}_{sc}$ over $\mathcal{L}_{c}$ in the zero-shot anomaly detection setting.\\
\indent Back to Tab.~\ref{table:ablation}, the result in the fifth row demonstrates that $\mathcal{L}_{sm}$ enhances the overall performance by guiding the model’s attention toward more challenging regions, thereby unlocking its optimal performance. Fig.~\ref{fig:ablation_lsm} visually illustrates the impact of $\mathcal{L}_{sm}$, showing that it more effectively separates the anomaly score distributions of normal and abnormal pixels.\\
\indent The penultimate row of Tab.~\ref{table:ablation} indicates that \textbf{\textit{``RL''}} significantly improves the model's localization performance by amplifying the residual differences between normal and abnormal regions. Specifically, it increases the pixel-level scores by \textcolor{purple}{0.25$\uparrow$}/\textcolor{purple}{8.55$\uparrow$}/\textcolor{purple}{4.36$\uparrow$}/\textcolor{purple}{1.03$\uparrow$} on MVTec-AD and \textcolor{purple}{0.02$\uparrow$}/\textcolor{purple}{5.18$\uparrow$}/\textcolor{purple}{3.52$\uparrow$}/\textcolor{purple}{0.31$\uparrow$} on VisA. Fig.~\ref{fig:ablation_rl} further visualizes the effectiveness of \textbf{\textit{``RL''}}, demonstrating that it enables more precise localization while effectively suppressing background noise.\\
\indent {A key aspect of residual learning is the application of a stop-gradient operation to the feature residuals, which ensures that only the segmentation head is updated. This design prevents the reconstruction model from being biased by pseudo anomalies. The comparison between the last two rows of Tab.~\ref{table:ablation} further substantiates the importance of this mechanism. Notably, removing the stop-gradient operation results in a substantial decline in pixel-level performance, primarily due to the bias introduced into the reconstruction model by pseudo anomalies.}

\subsubsection{Hyperparameter Analysis}
\label{sec-hyperparameterablation}
As shown in Fig.~\ref{fig:ablation-hyber}, we conduct comprehensive hyperparameter ablation studies on the MVTec-AD~\cite{MVTec-AD} dataset, examining the effects of the number of INPs $M$, the temperature parameter $\gamma$ in $\mathcal{L}_{sm}$, and the loss weight $\lambda$. Fig.~\ref{fig:ablation-hyber}(a) illustrates the impact of the hyperparameter $M$ on model performance. Experimental results indicate that image-level performance tends to stabilize when $M$ exceeds 4, while pixel-level performance continues to improve as $M$ increases. This is attributed to the fact that a larger number of INPs facilitates more detailed feature reconstruction, thereby enhancing the overall detection accuracy. However, an excessively large $M$ also leads to increased computational cost. Therefore, $M$ is set to 6 in our study to strike a balance between performance and efficiency. Fig.~\ref{fig:ablation-hyber}(b) shows the impact of the hyperparameter $\gamma$ on model performance. As $\gamma$ increases, pixel-level performance initially improves but subsequently declines. This is because a small $\gamma$ limits $\mathcal{L}_{sm}$'s ability to focus on challenging regions, while an excessively large $\gamma$ causes it to overlook the optimization of easy regions. In our study, $\gamma$ is set to 3. Fig.~\ref{fig:ablation-hyber}(c) illustrates the effect of the hyperparameter $\lambda$ on model performance. The results indicate that our method is not sensitive to $\lambda$. In our study, $\lambda$ is set to 0.2.

\begin{table}[]
\centering
\caption{Influence of the \textbf{Input Size} on model performance for the MVTec-AD~\cite{MVTec-AD} dataset. R256$^2$-C224$^2$ denotes resizing the image to 256×256, followed by a center crop to 224×224.}
\vspace{-10pt}
\fontsize{10}{14}\selectfont{
\resizebox{\linewidth}{!}{
\begin{tabular}{c|cc|c}
\toprule[1.5pt]
Input Size~$\downarrow$ & Image-level       & Pixel-level             & FLOPs(G)  \\ \midrule
R256$^2$-C224$^2$    & 99.4/99.8/99.0 & 98.6/70.2/68.1/95.2 & \textbf{34.7}   \\
R320$^2$-C280$^2$     & 99.7/\textbf{99.9}/\textbf{99.3} & 98.6/72.9/70.0/95.8 & 54.0 \\
\rowcolor{Light} 
R448$^2$-C392$^2$    & \textbf{99.8}/\textbf{99.9}/\textbf{99.3} & \textbf{98.7}/\textbf{75.6}/\textbf{71.6}/\textbf{96.0} & 105.4 \\ \bottomrule[1.5pt]
\end{tabular}}}
\label{table-ablation_inputsize}
\end{table}

\subsubsection{Influence of Input Size}
Tab.~\ref{table-ablation_inputsize} presents the impact of input image size on model performance. As the input size increases, performance steadily improves. In this study, we adopt R448$^2$-C392$^2$ as the default resolution. Remarkably, our method still outperforms existing approaches even under the lower-resolution setting of R256$^2$-C224$^2$.
\begin{table}[]
\centering
\caption{Influence of the \textbf{ViT Architecture} on model performance for the MVTec-AD~\cite{MVTec-AD} dataset. We also report the model efficiency (\textbf{Params(M)/FLOPs(G)}).}
\vspace{-10pt}
\fontsize{10}{14}\selectfont{
\resizebox{\linewidth}{!}{
\begin{tabular}{c|cc|c}
\toprule[1.5pt]
Architecture~$\downarrow$ & Image-level       & Pixel-level             & Efficiency  \\ \midrule
ViT-Small    & 99.5/99.8/98.8 & 98.5/74.7/70.9/95.5 & \textbf{37.1}/\textbf{31.3}   \\
\rowcolor{Light} 
ViT-Base     & \textbf{99.8}/99.9/99.3 & 98.7/75.6/71.6/96.0 & 142.7/105.4 \\
ViT-Large    & \textbf{99.8}/\textbf{100.0}/\textbf{99.5} & \textbf{98.9}/\textbf{77.0}/\textbf{72.8}/\textbf{96.6} & 365.2/271.3 \\ \bottomrule[1.5pt]
\end{tabular}}}
\label{table_ablation_vit_arch}
\end{table}
\subsubsection{Influence of ViT Architecture}
Tab.~\ref{table_ablation_vit_arch} illustrates the impact of the ViT architecture on model performance. Our method achieves strong detection results even with ViT-Small, with performance further improving as the ViT model size increases. Although ViT-Large achieves the best overall performance, its substantial computational cost and parameter count limit its practicality. Consequently, we adopt ViT-Base as the default architecture in this study.

\begin{table}[!t]
\centering
\caption{\textbf{Comparison of computational efficiency among SOTA methods}. \textbf{mAD} represents the average value of seven metrics on the Real-IAD~\cite{real-iad} dataset. The \textbf{INP-Former++-S} denotes a model variant based on the ViT-Small architecture, while \textbf{INP-Former++-S$^*$} refers to a model variant using the ViT-Small architexture with an image size of R256$^2$-C224$^2$.}
\vspace{-10pt}
\fontsize{10}{14}\selectfont{
\resizebox{0.9\linewidth}{!}{
\begin{tabular}{c|ccc}
\toprule[1.5pt]
Method~$\downarrow$              & Params(M)     & FLOPs(G)     & mAD           \\ \midrule
RD4AD~\cite{RD4AD}               & 150.6          & 38.9         & 68.6          \\
UniAD~\cite{uniad}               & \textbf{24.5} & \textbf{3.6} & 67.5          \\
SimpleNet~\cite{liu2023simplenet}           & 72.8          & 16.1         & 42.3          \\
DeSTSeg~\cite{zhang2023destseg}             & 35.2          & 122.7        & 64.2          \\
DiAD~\cite{diad}                & 1331.3        & 451.5        & 52.6          \\
MambaAD~\cite{he2024mambaad}             & \underline{25.7}          & \underline{8.3}          & 72.7          \\
Dinomaly\footref{reproduce}~\cite{guo2024dinomaly}            & 132.8         & 104.7        & 76.3            \\ 
INP-Former\footref{reproduce}~\cite{luo2025exploring}            & 139.8         & 98.0        & 78.0            \\ \midrule
\rowcolor{Light} 
\textbf{INP-Former++} & 142.7         & 105.4           & \textbf{79.4} \\ 
\textbf{INP-Former++-S} & 37.1         & 31.3           & \underline{79.1} \\ 
\textbf{INP-Former++-S$^{*}$} & 37.1        & 10.3          & 76.3 \\ 
\bottomrule[1.5pt]
\end{tabular}}}
\label{table:complexity}
\end{table}
\subsection{Complexity Comparisons}
\label{sec-complex}
Tab.~\ref{table:complexity} compares the proposed INP-Former++ with eight SOTA methods, including our conference version, INP-Former, in terms of model performance and computational complexity. With comparable FLOPs, our INP-Former++ outperforms Dinomaly by \textcolor{purple}{3.1$\uparrow$}. Compared with our earlier version, INP-Former, the proposed INP-Former++ introduces only an additional 2.9M parameters and 7.4G FLOPs, yet achieves a \textcolor{purple}{1.4$\uparrow$} performance gain. We also report the efficiency and performance of two variants: INP-Former++-S and INP-Former++-S*. INP-Former++-S substantially reduces both the parameter count and computational cost, with only a minor performance drop of \textcolor{purple}{0.3$\downarrow$}. INP-Former++-S* achieves a performance improvement of \textcolor{purple}{3.6$\uparrow$} while maintaining comparable FLOPs to MambaAD. Overall, our method demonstrates strong potential for industrial applications.

\subsection{Exploration on INPs}
\begin{figure}
    \centering
    \includegraphics[width=\linewidth]{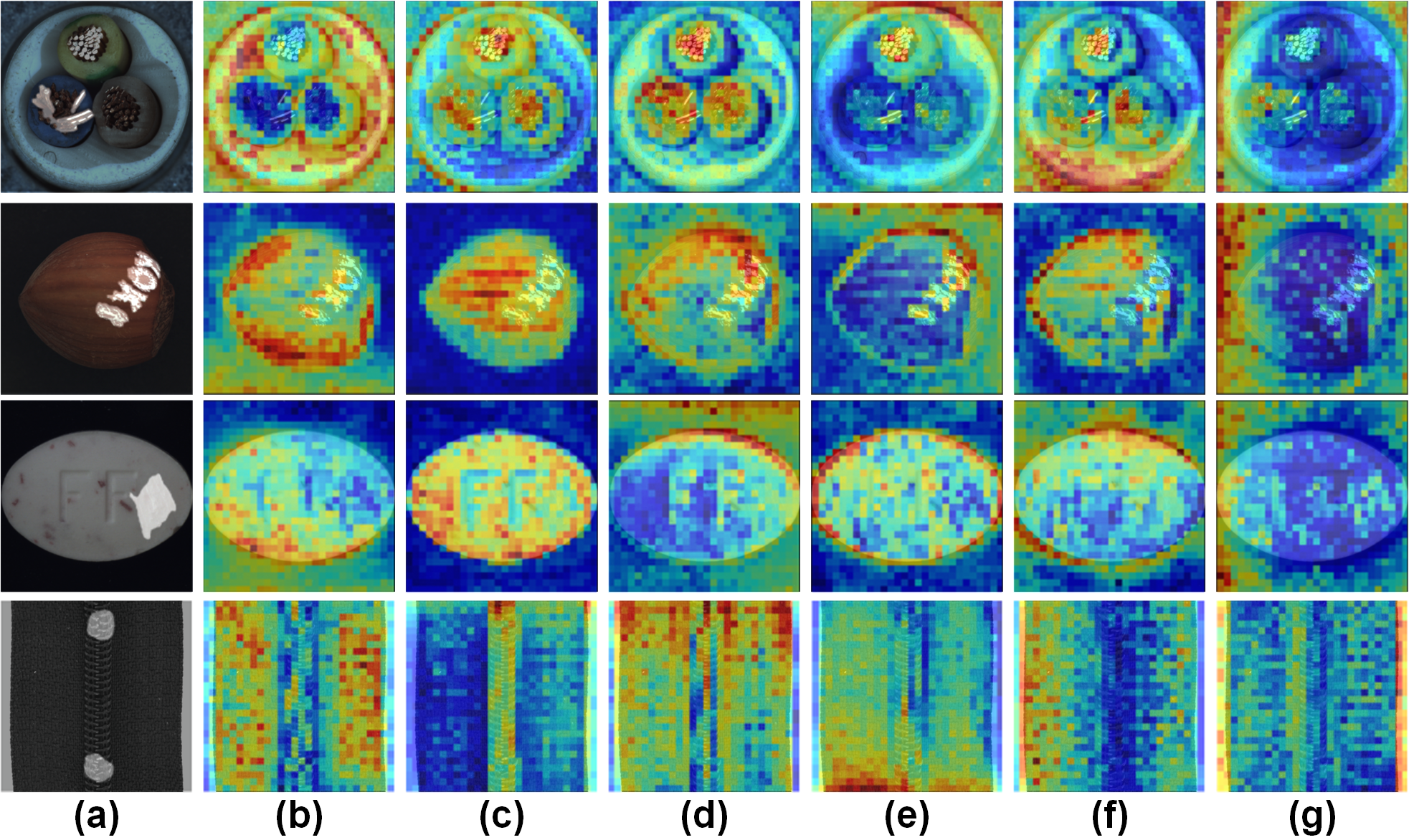}
    \vspace{-20pt}
    \caption{\textbf{Visualizations of INPs.} (a) Input anomalous image and
ground truth. (b)-(g) Cross attention maps between INPs and image
 patches.}
    \label{fig:prototypevis}
\end{figure}
\subsubsection{Visualizations of INPs}
Fig.~\ref{fig:prototypevis} illustrates that INP effectively captures diverse semantic representations within the image, including object regions (Fig.~\ref{fig:prototypevis}(b), (c) and (d)), object boundaries (Fig.~\ref{fig:prototypevis}(e) and (f)), and background areas (Fig.~\ref{fig:prototypevis}(g)). This diversity is attributed to our design of guiding the reconstruction process with INPs. Furthermore, the Soft INP Coherence Loss $\mathcal{L}_{sc}$ encourages the INP to consistently capture normal features while avoiding the capture of anomalous ones during testing. This mechanism ensures that the decoded features contain only normal information, thereby enhancing anomaly detection performance.

\begin{figure}
    \centering
    \includegraphics[width=\linewidth]{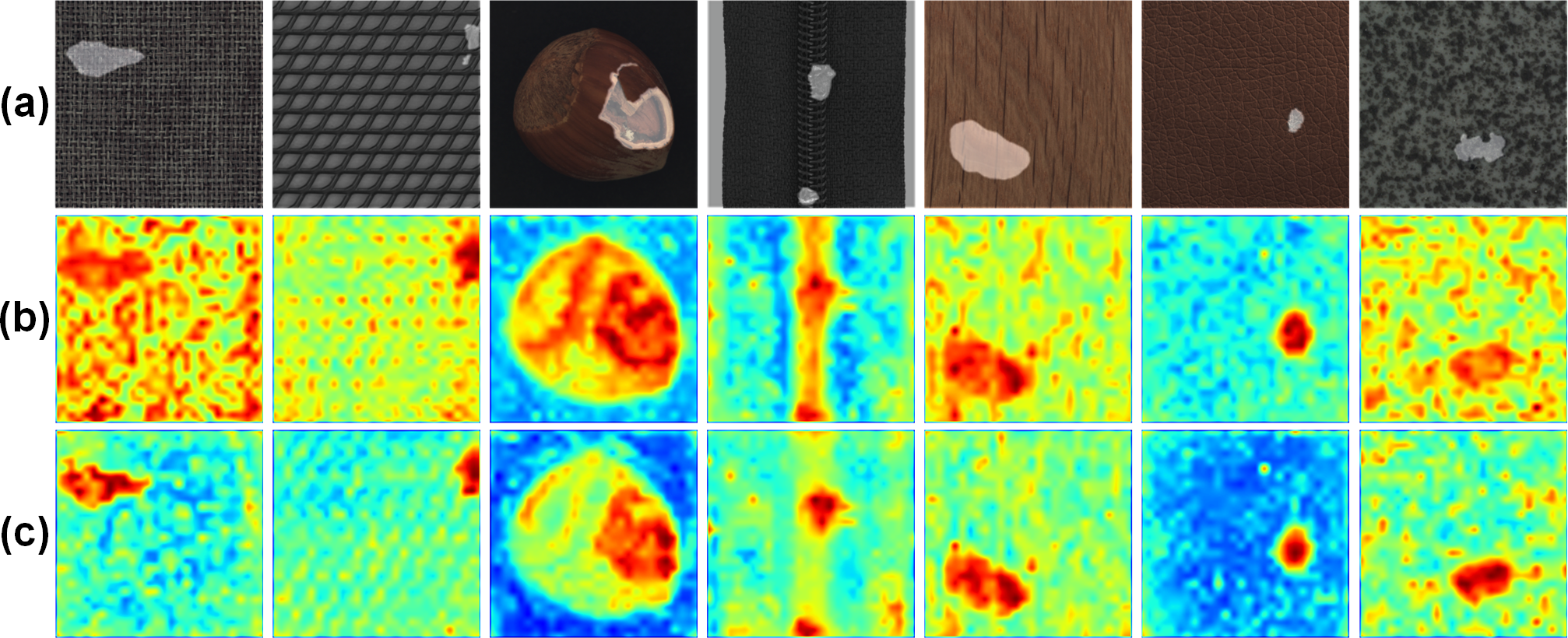}
    \vspace{-20pt}
    \caption{\textbf{Zero-shot anomaly detection results}. (a) Input anomalous image and ground truth. (b) Anomaly map from \textbf{INP-Former (w/ $\mathcal{L}_{c}$)}. (c) Anomaly map from \textbf{INP-Former++ (w/ $\mathcal{L}_{sc}$)}. Both INP-Former and INP-Former++ are trained on the Real-IAD~\cite{real-iad} dataset and evaluated on MVTec-AD~\cite{MVTec-AD}. The distance map is upsampled to the original image resolution to generate the final anomaly map.}
    \label{fig:anomalymapZS}
\end{figure}
\begin{table}[!t]
\centering
\caption{\textbf{Zero-shot} anomaly detection performance on \textbf{MVTec-AD}~\cite{MVTec-AD} dataset.}
\label{table:zero-shot}
\vspace{-10pt}
\fontsize{10.5}{14}\selectfont{
\resizebox{0.95\linewidth}{!}{
\begin{tabular}{c|cc}
\toprule[1.5pt]
Method~$\downarrow$       & Image-level    & Pixel-level         \\ \midrule
APRIL-GAN~\cite{aprilgan}        & \underline{86.1}/\underline{93.5}/\underline{90.4} & \underline{87.6}/\textbf{40.8}/\textbf{43.3}/{44.0} \\
WinCLIP~\cite{WinClip}        & \textbf{91.8}/\textbf{96.5}/\textbf{92.9} & 85.1/-/31.7/64.6 \\ \midrule
\textbf{INP-Former\footref{reproduce} (w/ $\mathcal{L}_{c}$)}~\cite{luo2025exploring} & 80.8/90.7/89.1 & \textbf{87.9}/34.2/38.1/\underline{77.3} \\
\rowcolor{Light}
\textbf{INP-Former++ (w/ $\mathcal{L}_{sc}$)} & 81.9/{90.4}/{90.0} & \textbf{87.9}/\underline{34.9}/\underline{38.8}/\textbf{79.6} \\ \bottomrule[1.5pt]
\end{tabular}}}
\end{table}

\subsubsection{Generalization capabilities of INP Extractor}
As illustrated in Fig.~\ref{fig:anomalymapZS}, the INP Extractor trained on the RealIAD~\cite{real-iad} dataset exhibits strong generalization capability to the unseen MVTec-AD~\cite{MVTec-AD} dataset, where the distance maps between feature tokens and the extracted INPs can be effectively utilized for zero-shot anomaly detection. These results clearly validate the INP Extractor’s ability to dynamically extract representative INPs from a single image. Furthermore, the quantitative results in the zero-shot setting, as shown in Tab.~\ref{table:zero-shot}, demonstrate that $\mathcal{L}_{sc}$ enhances the model's detection and localization performance compared to $\mathcal{L}_{c}$. The comparison between (b) and (c) in Fig.~\ref{fig:anomalymapZS} clearly highlights the superiority of $\mathcal{L}_{sc}$ over $\mathcal{L}_{c}$. In comparison to methods specifically designed for zero-shot anomaly detection, such as APRIL-GAN~\cite{aprilgan} and WinCLIP~\cite{WinClip}, our INP-Former++ achieves the best AUPRO result, reaching \textbf{79.6}.

\begin{figure}[t]
    \centering
    \includegraphics[width=0.98\linewidth]{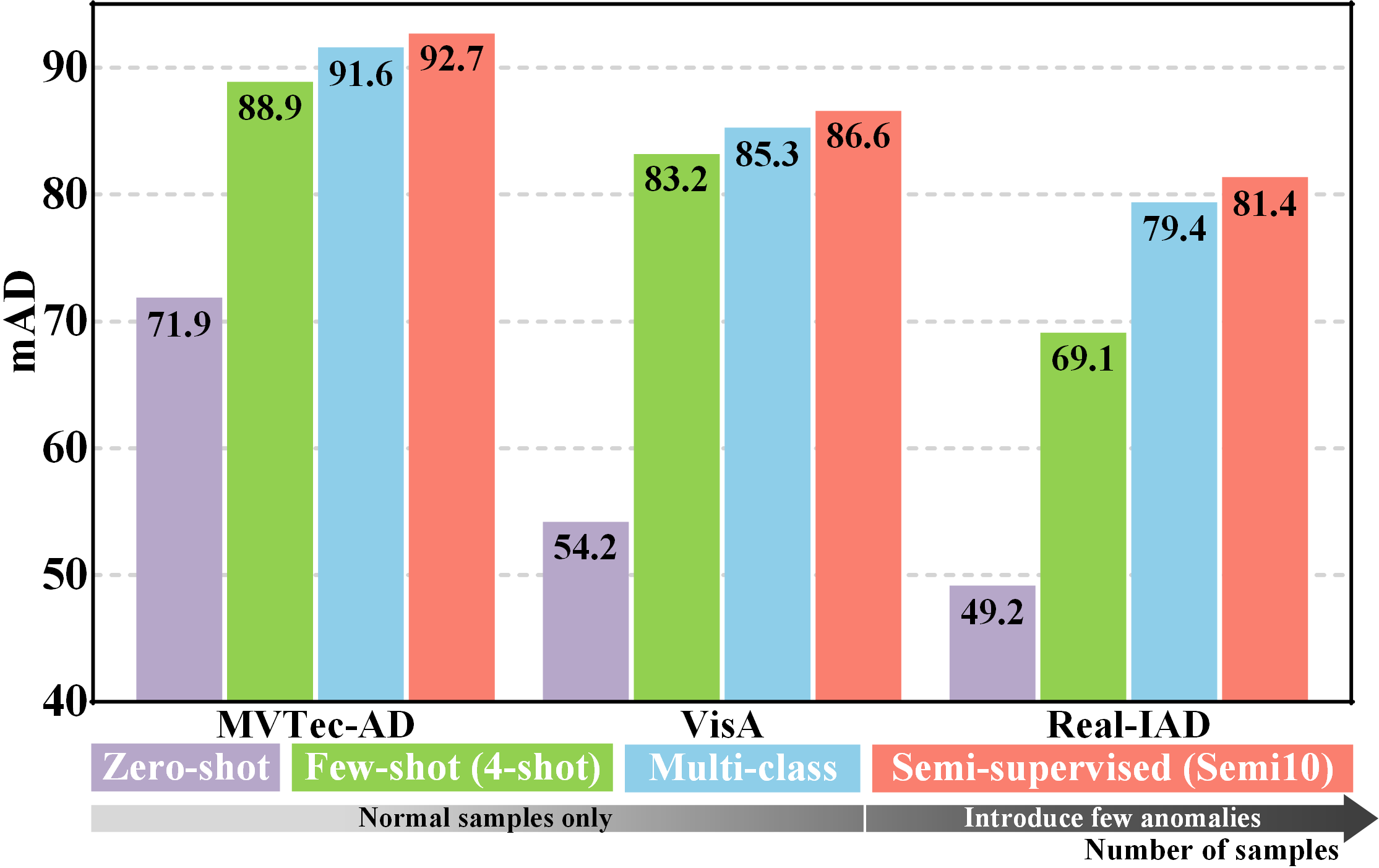}
    \vspace{-12pt}
    \caption{\textbf{Performance evaluation of INP-Former++ under different settings.} mAD represents the average value of seven metrics.}
    \label{fig:scale}
\end{figure}
\subsection{Scalability of INP-Former++}
Fig.~\ref{fig:scale} illustrates the overall performance of INP-Former++ across three datasets under different settings. The experimental results highlight the strong scalability of our method, with performance steadily improving as the number of normal samples increases, progressing from the zero-shot setting to the few-shot setting, and then to the multi-class setting. Subsequently, with the introduction of a small number of anomalies, the performance further improves in the semi-supervised setting. This scalability demonstrates significant potential for real-world industrial applications.

\section{Conclusion}
In this work, we extend our original INP-Former to INP-Former++, a novel framework for universal anomaly detection that systematically explores the role of INPs. INP-Former++ learns to aggregate the semantics of normal tokens into INPs, which are then leveraged to guide the reconstruction of normal tokens. By introducing residual learning, the framework amplifies the residual differences between normal and anomalous regions, significantly boosting anomaly detection performance. The proposed Soft INP Coherence Loss and Soft Mining Loss further enhance the quality of INPs and optimize the training process. Extensive experiments on both industrial and medical datasets demonstrate that INP-Former++ achieves SOTA performance across single-class, multi-class, few-shot, and semi-supervised anomaly detection settings. These results confirm the existence and effectiveness of INPs, which can be reliably extracted even from images of unseen categories, thereby enabling zero-shot anomaly detection. {However, the zero-shot performance of our method is currently suboptimal, which we attribute to the lack of language prompts. In future work, we aim to incorporate language prompts to elevate the zero-shot performance of our method to SOTA levels, thereby further advancing the boundary of universal anomaly detection.}

\ifCLASSOPTIONcaptionsoff
  \newpage
\fi



\bibliographystyle{IEEEtran}
\bibliography{IEEEabrv,ref.bib}

\end{document}